\RequirePackage{fix-cm}
\RequirePackage{rotating}
\documentclass[smallextended, natbib]{svjour3}       % onecolumn (second format)
\smartqed  % flush right qed marks, e.g. at end of proof
\usepackage{graphicx}
\usepackage[linesnumbered,ruled]{algorithm2e} % For algorithms
\usepackage{booktabs} % For formal tables
\usepackage{placeins} %for better control of tables
\usepackage{tabulary}
\usepackage[T1]{fontenc}
\usepackage{url}
\usepackage{amsmath}
\usepackage{rotating}
\usepackage{hyperref}
\makeatletter
\newcommand*\bigcdot{\mathpalette\bigcdot@{.5}}
\newcommand*\bigcdot@[2]{\mathbin{\vcenter{\hbox{\scalebox{#2}{$\m@th#1\bullet$}}}}}
\makeatother

\journalname{Preprint}
\begin{document}

\title{Creating user stereotypes for persona development from qualitative data through semi-automatic subspace clustering
%\thanks{This study was part of the ELDORADO project ``Preventing malnourishment and promoting well-being among the older adults at home through personalised cost-effective food and meal supply'' supported by a grant (4105-00009B) from the Innovation Fund Denmark.}
}

\titlerunning{Creating user stereotypes for persona development}        % if too long for running head

\author{Dannie Korsgaard \and Thomas Bj\o rner \and Pernille Krog S\o rensen \and Paolo Burelli
}

\authorrunning{Korsgaard, Bj\o rner, S\o rensen, and Burelli} % if too long for running head

\institute{D. Korsgaard \and T. Bj\o rner \and P. K. S\o rensen \at Aalborg University Copenhagen, A. C. Meyers V{\ae}nge 15, 2450, Copenhagen SV, Denmark \\
Tel.: +45 90 40 26 24\\
\email{dmk@create.aau.dk} \\
\and
P. Burelli \at IT University of Copenhagen, Rued Langgaards Vej 7, 2300 Copenhagen, Denmark \\
}
\date{Received: date / Accepted: date}

\maketitle

\begin{abstract}
Personas are models of users that incorporate motivations, wishes and objectives; These models are employed in user-centred design to help design better user experiences and have recently been employed in adaptive systems to help tailor the personalised user experience.
Designing with personas involves the production of descriptions of fictitious users, which are often based on data from real users. The majority of data-driven persona development performed today is based on qualitative data from a limited set of interviewees and transformed into personas using labour-intensive manual techniques. In this study, we propose a method that employs the modelling of user stereotypes to automate part of the persona creation process and addresses the drawbacks of the existing semi-automated methods for persona development. The description of the method is accompanied by an empirical comparison with a manual technique and a semi-automated alternative (multiple correspondence analysis). The results of the comparison show that manual techniques differ between human persona designers leading to different results. The proposed algorithm provides similar results based on parameter input, but was more rigorous and will find optimal clusters, while lowering the labour associated with finding the clusters in the dataset. The output of the method also represents the largest variances in the dataset identified by the multiple correspondence analysis.

\end{abstract}

\section{Introduction}

In software development, projects have a reputation for failure \citep{Savolainen2012}, and this reputation has previously been attributed to the lack of user involvement \citep{Ewusi-Mensah2003}. To actively keep users in focus during the software design process (referred to as ``design time'' by \cite{Fischer2001}), the use of personas has been encouraged as a tool for modelling users of software systems in a readily comprehensible manner understandable to any member of the design team \citep{Cooper1999,Pruitt2006,Pruitt2003}. 

Personas were popularized by \cite{Cooper1999} as research-based fictional archetypes modelled from real users. Personas are descriptions of the target group(s) for the product under development, and they are created to avoid what \cite{Cooper2007} refers to as \textit{the elastic user} in situations where user trials are unavailable. The elastic user is an abstract concept of a user, for whom every developer will have a different mental impression and assign different needs. Describing real users by a persona with a name and a list of needs will challenge these mental impressions and make assumptions more explicit \citep{Miaskiewicz2011}. 

In recent years a few examples have investigated the application of personas beyond traditional user-centred design and within the context of user personalisation as a companion tool to classical user modelling~\citep{Madureira2014,Holmgard2014}. The primary argument for this is that better personalisation can be achieved by modelling user motivations~\citep{Melhart2019}. In some contexts, such as computer games in which the decisions are driven by the game rules~\citep{Holmgard2014}, the motivations can be extracted from user behaviour data; in other cases, personas have been shown to be an effective tool to interpret the users' decision-making process and to shape more effectively the personalisation~\citep{Casas2008,Madureira2014}.

Recent investigations of how personas are currently developed by practitioners have shown that most practitioners build personas from a qualitative data foundation collected through ethnographic studies and interviews \citep{Viana2016, Nielsen2014, Brickey2012}. The challenges involved in the development of personas or user models from qualitative data are two-fold: on one side, the manual techniques applied to aggregate qualitative data have generally been criticized as consuming excessive resources, not being applicable to large data amounts, lacking rigour, and relying on subjective judgments \citep{Brickey2012, Macia2015, Guest2003}.
On the other side, algorithmic methods commonly employed in user modelling on quantitative data are efficient in these areas, as they are executed with rigour by a computer at high speeds; however, the automated and semi-automated techniques in persona development have been criticized for being often overly complex, giving a false appearance of precision and to inspire non-critical acceptance of results \citep{Siegel2010}.

% The sophisticated algorithms contained in some of these techniques can potentially undermine the transparency of how the data are turned into personas among members of the development team who lack the theoretical foundation for understanding how the algorithms work; thus, threatening the believability and adoption of the personas among members of the development team \citep{Viana2016}.

One of the primary problems with most algorithms for user data segmentation is that they tend to be most effective when applied to datasets with low dimensionality and high cardinality (i.e., a low number of features and a high number of data points) and may be problematic to use with data from a small number of interviewees. Viana and Robert \citep{Viana2016} found that practitioners used data from an average of twenty users as the foundation for persona development. Data from this amount of users are not unusual in qualitative studies due to the high volume of data collected per subject and the manual work associated with data analysis \citep{Bjrner2015}. \cite{Moser2012} recommended a sample size of at least 2\textsuperscript{\textit{k}} before semi-automated clustering methods can be employed for persona development.
To elaborate, \textit{k} is the number of features (e.g., demographics, tasks, attitudes, preferences, needs) in which the users vary, and along which the users are divided into segments. With data from twenty users, this equation will result in the limiting amount of four features. Practitioners have been found to include information in their personas related to demographics, job responsibilities, tasks and skills, tools usage patterns, and big pain points \citep{TaraMatthewsTejinderJudge2012, Viana2016}. As the amount of information included in a persona may vary from project to project and from persona designer to persona designer (PD), the method of analysis should strive to accommodate data with few samples, including many features (referred to as sparse data).

%IMPORTANT: find something about ethnographic data/interview data and missing data - we need to be able to handle missing data!!

In this study, we aim to design and evaluate a user modelling method that applies to qualitative data and can be used for persona development, while being understandable to lay members of the development team. Our study is based on interview transcript data obtained from a qualitative study in connection with the ELDORADO project \citep{CopenhagenU2018}. In the ELDORADO project, we are concerned with the design of diet-related information and communication technology for older members of the Danish population. The aim is to use information and communication technology to promote healthy food habits through cost-efficient digital solutions and personalized interventions.

\section{Related work}

\subsection{The Goodwin Method}

Currently, only a few methods \citep{Laporte2012,Miaskiewicz2008} use qualitative interview data for persona development and they appear to have the following common drawbacks:

\begin{itemize}
  \item High dependency on well-structured interview questions.
  \item Use of complex dimensionality reduction techniques, hampering the dissemination of information regarding the development process.
\end{itemize}

Firstly, the methods rely on the use of structured questions to facilitate systematic comparisons between answers. To create this set of questions, the PD must have prior knowledge of what the persona may contain. The ideal method for persona development needs to be compatible with exploratory methods, which employ a vague questioning structure to expand the learning outcome beyond the preconceptions of the PD. When providing the interviewer more freedom, the potential for irrelevant or missing information rises. Thus, a suitable method should be used to locate relevant information and deal with missing data. Secondly, methods relying on dimensionality reduction techniques, such as singular value decomposition (SVD), should be avoided to allow for better transparency in how the data are turned into the final personas. To compile an easy-to-comprehend persona development method, inspiration was sought in an existing, well-established persona development method with proven didactic properties.

Cooper has, in the wake of the popularity he brought to the persona design method, made a business of consulting and educating Human-Computer Interaction professionals \citep{Cooper2018}. Cooper's methods have since been developed by Goodwin, who filled the role of vice president and general manager in Cooper's company and, through several iterations, turned the methods into an interaction design curriculum \citep{Adlin2006}. Goodwin's method for designing personas is performed manually and has been described in several books and articles \citep{Goodwin2009,Cooper2007,Goodwin2002} and is widely used \citep{Kerr2014a, Calde2002, Antle2006, Christidis2011}. Her method (as described in \citep{Cooper2007}) outlines the following six necessary stages for moulding qualitative data into final personas:
% enumerate
\begin{enumerate}
\item Identifying behavioural variables
\item Mapping interview subjects to behavioural variables
\item Identifying significant behavioural patterns
\item Synthesizing characteristics and relevant goals
\item Checking for redundancy and completeness
\item Expanding descriptions of attributes and behaviours
\end{enumerate}
In stages one and two, the interview data are quantified to numeric values by mapping subjects to identified behavioural variables using visual analogue scale (VAS). Each VAS represents a behavioural dimension and all of the VASs are combined to create a multidimensional space where each subject's behaviour is mapped by a data point. After each subject's behaviour has been quantified in terms of markings in a set of VAS, the behavioural patterns are identified through a segmentation process in stage three, which is described as follows\citep{Cooper2007}:
% quote
\begin{quote}
``$\dotsc$ look for clusters of subjects that occur across multiple ranges or [behavioural] variables. A set of subjects who cluster in six to eight different variables (changed to one-third of the variables in \citep{Goodwin2009}) will likely represent a significant behaviour pattern that will form the basis of a persona.''\end{quote}

The automation of this manual segmentation through user modelling could not only speed up the process and remove workload from the PD but potentially improve the resulting segmentation by finding \textit{optimal clusters}. To understand what is meant by optimal clusters, it should be highlighted that the process employed for finding clusters across behavioural variables involves a tradeoff between group size and the number of behavioural variables that the group members have in common \citep{Chapman2008}. For example, we may consider the extremes: If features are compared across every individual in the population of Europe, very few features would be shared (e.g., ``living in Europe'' and ``being human''). This is a significant pattern (as many people can share it), but it lacks details to inform a design process. The comparison of two arbitrary people in Europe is likely to yield a larger range of shared features (e.g., they may enjoy the same hobby, prefer the same cereal brand, dislike horror movies, etc., as well as ``being human'' and ``living in Europe''). The long list of features will be specific to the two individuals, and it is unlikely to represent any other citizens in Europe; thus, it is considered to be a less significant pattern. Goodwin's method utilizes a number (e.g., six to eight, or one-third) of the shared features as a rule-of-thumb to determine when a significant pattern has been found. However, instead of relying on a rule-of-thumb, we consider a successful segmentation into significant behavioural patterns to obtain optimal clusters that group the most subjects across as many features as possible. This is an optimization problem where computers are often used to assist in finding solutions as the number of subjects and variables increases. 

%The use of manual qualitative techniques for segmenting users is common in the area of persona development, although these methods have generally been criticized as consuming excessive resources and relying on subjective judgments \citep{Brickey2012}. By contrast, other techniques adopted from market segmentation seek to develop personas based on quantitative analysis and collection techniques. Examples of these techniques include principal component analysis \citep{Sinha2003}, exploratory factor analysis \citep{McGinn2008}, and cluster analysis \citep{Wockl2012}. A layering approach has also been suggested where segmentation is performed based on large collections of quantitative data, which transforms the segments into relatable characters with details obtained from qualitative interviews \citep{Marcengo2009, McGinn2008, Pruitt2003}.

\subsection{Segmenting qualitative data}

One key step in the aforementioned personas development method requires to operationalise and segment qualitative data such as open-ended interviews; however, only a few studies have proposed methods using algorithmic quantitative techniques on data collected using qualitative methods. \cite{Laporte2012} collected open-ended textual responses regarding children's technology usage, then coded the data into categories and performed a multiple correspondence analysis (MCA) on the category scores to guide a manual segmentation process. \cite{Miaskiewicz2008} conducted a latent semantic analysis (LSA) using interview transcripts to calculate the similarity scores between answers and then employed clustering algorithms with the similarity scores to create groups. Once the data were segmented and the groups created, features were extracted from the groups and used to produce final persona descriptions.

The methods applied by both \cite{Laporte2012} and \cite{Miaskiewicz2008} contained two phases: (1) quantifying text into numeric or categorical data, (2) pattern detection, and data segmentation (not always separable, thus treated as one phase).
The output of the second step can be described as intermediate user models, which are used as the basis for creating personas.
In the following sections, a more detailed description is given of relevant work concerned with each of these phases.

\subsubsection{Quantifying text into numeric or categorical data}

%talk about how Miaskiewicz2008 and LSA uses frequency data (counts) to make text into numeric data
%problem with LSA is also missing data?!
Qualitative data often comes in the form of interviews, videos, drawings, or other media not directly compatible with semi-automated methods \citep{Bjrner2015}. One exception is the study by \cite{Miaskiewicz2008}, in which qualitative textual interview data were translated into quantitative numeric data purely by computing the word counts of each textual answer to a structured set of questions. Thus, the information about in which order the words appeared in the textual answers was discarded. A preprocessing step may occur before word counts are established. The process referred to as stemming fuses words that are variations of the same stem, thus reducing the count of words into reduced counts of word stems \citep{Grimmer2013}. To avoid that stop words (common words of a language, such as conjunctions) dominate the analysis, each word count is weighted by how rare the word is in the total body of text (also called term frequency by inverse document frequency weighting) \citep{Grimmer2013}. While the conversion of text to numeric data is automated with LSA, certain limitations are linked to this approach. The output is a list of word frequency counts weighted by word rarity for each answered question. Using frequency counts is justified by the distributional hypothesis (words that appear in the same contexts tend to convey similar meanings) \citep{Sahlgren2008}, but the approach struggles in several cases, such as in irony, metaphors, and words with multiple meanings. Besides, the quantification of the textual responses requires a structured set of questions, as each question results in an answer that functions as a unit for comparison. Texts from ethnographic interviews are not guaranteed to have this level of structure.

\cite{Laporte2012} used a manual approach instead, where answers to a structured set of questions were turned into categorical data, such that similar answers to each question were placed in the same category. The output of the process is a list of categorical variables for each question, and the process leaves the interpretation and categorization of answers to the person performing the conversion. Another manual approach to quantifying qualitative data into numeric data for persona development was proposed by \cite{Goodwin2002}. Here, behavioural variables are identified manually in the textual data and outlined as VASs, defined by two extremes. Interviewees are then manually mapped on to the scales in accordance with their recorded behaviour. By treating each VAS as a number between zero and one, this process converts qualitative data into quantitative data at the interval level \citep{Stevens1946}. The manual methods are prone to bias, but at present, manually organizing and transforming the text into quantitative data may be the best way to deal with semi- or unstructured textual data.

%Although the purpose of a study by Guest and Mclellan \citep{Guest2003} was to do thematic analysis and not to develop personas, their technique could just as well have been applied to persona development. Guest and Mclellan coded interview transcripts ... \citep{Guest2003} \citep{MacQueen} \citep{Macia2015}

%Option 3 would be better in cases where compare you have text that e.g. talks about something to prove that something else is better: "At have de der lækre ting gør ikke børneværelset sjovere at være på end andre billigere ting"

%also add goodwin's method here with manual mapping in a dataspace.

\subsubsection{Pattern detection, data segmentation and user modelling}
%LSA good understanding by using wiki (https://en.wikipedia.org/wiki/Latent_semantic_analysis) and this video on SVD (https://www.youtube.com/watch?v=P5mlg91as1c)

Once the textual data are converted either manually or automatically to numeric or categorical data, a set of quantitative methods can be applied to detect patterns, segment users, and create user models. Common techniques include dimensionality reduction and/or cluster analysis.

Dimensionality reduction is a technique for transforming data expressed by a set of variables into a smaller set of variables that ideally account for the observed properties of the data \citep{VanderMaaten2007}. \cite{Sinha2003} was the first to propose the use of dimensionality reduction for persona development by utilizing principal component analysis. Numeric data represented by 32 dimensions were given as input to the principal component analysis to find a smaller set (5) of new orthogonal dimensions that were rotated to fit the most variance in the data set. By using equamax rotation variances were distributed more uniformly across the found dimensions, and the set of new dimensions were then used as inspiration for writing the persona descriptions \citep{Sinha2003}.

Both of the methods for persona development proposed by \cite{Miaskiewicz2008} and \cite{Laporte2012} uses dimensionality reduction through the concept of SVD. SVD is a process that, through a set of linear algebra operations, reduces a matrix of observations and variables into a product of three matrices describing the relationship between observations (rows) and variables (columns) of the original matrix in terms of a set of concepts \citep{Leskovec2014}. Each of the outputted matrices describe, respectively, the observations-to-concept relationships and the variables-to-concept relationships, as well as the strength of each concept, and by sorting the concepts according to strength, the least impactful concepts can be discarded to, for example, separate patterns from noise, compress the data, or visualize the most important trends in a data set \citep{Leskovec2014}.

Particularly, LSA uses SVD to produce concepts that, in terms of word occurrences, represents common answers \citep{Miaskiewicz2008}. When performing LSA on a large body of text, general semantic spaces can be extracted and used for text comparisons \citep{Landauer1998}. To compare textual answers from the interviewees, \cite{Miaskiewicz2008} used the common tasaALL semantic space, which was built from more than 92409 words and reduced to 419 dimensions and contains texts used in the curriculum up to the college level \citep{Landauer2007}. The average similarity scores across all questions of two interviewees were then used as input to an average-linked hierarchical clustering algorithm that produces a dendrogram visualizing the similarity between interviewees. The segmentation was done by cutting the resulting dendrogram when all interviewees were grouped with at least one other interviewee \citep{Miaskiewicz2008}. The relevance and the number of texts used to generate the semantic space used for comparisons will impact the results of the generated personas. Using a space built from a small corpus of texts will only contain a limited amount of associations found in the few texts included, and if the texts used are irrelevant to the texts compared, irrelevant associations will have been made. Most of the available semantic spaces are developed from texts in major languages, such as English, French, and German. This makes the method less accessible for interviews done in a language where the availability of semantic spaces is low.

MCA takes categorical data expressed as binary indicators and displays similarities in the patterns of categories between interviewees as distances on a perceptual map. An intuition of how the distances are calculated between two interviewees can be made from the four following examples \citep{Husson2011}. (1) If they have been assigned the same categories, the distance between them will be zero. (2) If they have been assigned many of the same categories, the distance between them will be short. (3) If they have been assigned all the same categories except for one, which is assigned to one of the interviewees and rarely to all the other interviewees, they should be distanced to account for the uniqueness of one of them. (4) If they share a rare category, they should be placed close together, despite differences elsewhere, to account for their common distinctiveness. SVD is used as part of the MCA to create a low-dimensional space (called a perceptual map) while maintaining as much of the distances between interviewees as possible. The axes returned by the MCA are ordered by the amount of variance they portray, such that the first axis accounts for the most variance, while the last axis in the set will account for the least. By using the first axes in the returned set, the most significant differences among interviewees can be displayed on a 2D perceptual map. Through visual inspection of the map, a person may group the interviewees based on proximity, while being aware of the limitations of the map in terms of displayed variance. In the form proposed by \cite{Laporte2012} MCA-based personas appear to be a highly manual process where the computer-generated part simply serves as guidance, still leaving most of the labour to be performed by the PD.

%something about the problem of understanding
While dimensionality reduction is useful, the mathematics involved in the techniques are rather ``complex'' \citep{Sourial2010}, and the details are often left out in gentle introductions to, for example, correspondence analysis (MCA is a specialized version of correspondence analysis) \citep{Bock2017}. To comprehend the concept of SVD, an understanding of the fundamentals of linear algebra is required - a mathematical discipline that is commonly taught to undergraduates and has been reported to cause them some difficulty \citep{Grenier-Boley2014, Masters2000, Harel1987}. The output of dimensionality reduction is also counter-intuitive because the distances between data points can no longer be understood according to the original questions asked. These considerations make the use of dimensionality reduction techniques problematic to persona development, in which the process from data to the final persona appears to play an important role for persona acceptance \citep{Viana2016,Siegel2010}.

Clustering analysis is another approach for pattern detection and the segmentation of numeric data that is commonly used in user modelling~\citep{Burelli2015umuai}. This kind of analysis assumes non-uniformity in the data and seeks a structure by grouping data based on similarity. While a large number of clustering methods have been proposed \citep{Jain2010a}, a limited set of traditional clustering methods are commonly applied to create user models for persona development. As already described, hierarchical clustering worked as a supporting element in creating personas from the LSA \citep{Miaskiewicz2008}, but has also been used to group users directly by their click-streams \citep{Zhang2016} and survey answers \citep{Tu2010}. While hierarchical clustering builds a tree of nested interviewees by their similarity, another set of methods commonly used for persona development segments the interviewees into non-overlapping or overlapping clusters  \citep{Wockl2012,Masiero2011}. The process is performed in a multidimensional data space composed of each recorded variable. This data space allows the algorithms to calculate a distance (representing similarity/dissimilarity) between the data points holding information about each interviewee. Partitional clustering methods assign each data point to a single cluster, such as the classical k-means algorithm \citep{Tan2005}, which does so based on the distances to cluster centroid. Fuzzy clustering methods assigns each datapoint a weighted membership to each cluster, such as the fuzzy c-means algorithm, which does so based on distance from each data point to the cluster centroid \citep{Bezdek1984}. Since qualitative studies tend to collect a high volume of data per subject, the data space will be characterized by having a high number of dimensions in relation to the low amount of data points. Conducting clustering in such a data space becomes problematic because of the phenomenon known as \textit{the curse of dimensionality}. The curse of dimensionality refers to the equilibrium of distances between data points that occurs when dimensions are added to form a higher dimensional data space. As more dimensions are added, the volume of the data space increases exponentially and spreads the data out to a point where the distances between them approach equality and clusters cease to exist. This is one of the reasons \cite{Moser2012} recommended that the sample size should grow exponentially with the number of dimensions before semi-automated clustering methods can be employed for persona development.

\subsection{Density-Based Optimal Projective Clustering}

To locate optimal clusters and extract the information needed to build the personas, we suggest using the density-based optimal projective clustering (DOC) algorithm \citep{Procopiuc2002} belonging to the branch of clustering algorithms known as subspace clustering \citep{Parsons2004, Muller2009}. Subspace clustering aims to find different combinations of dimensions where the data set clusters (the combination of dimensions is called a subspace). This is different from a traditional clustering approach that seeks to find clusters across ALL dimensions. Besides mitigating the effect of the curse of dimensionality, subspace clustering can ideally make more efficient use of the available data. When applied for persona development, subspace clustering takes into account that although two interviewees in a cluster might share a range of similarities, one of them may also have other things in common with a third interviewee in another cluster based on a set of alternative features. As will be elaborated on later, the DOC algorithm is a clustering method that utilizes randomized sampling to find optimal clusters, which to a high degree simulates what Goodwin's method dictates the PD to do manually. Like many other methods built around randomized sampling, DOC consists of a simple operation that is performed many times by the computer. Thus, even though a formal definition of DOC requires linear algebra, the simple mechanics of the method may be demonstrated to an audience manually by using a set of VASs, a ruler to measure distances, and a dice to illustrate random sampling. Besides the aforementioned contrast to dimensionality reduction techniques, the output of the DOC algorithm also remains intuitive because the original axes are retained.

The remainder of this paper will investigate the potential of a mixed-method for persona development using the DOC algorithm to replace stage three in Goodwin's method. Stage three produces what Goodwin refers to as \textit{proto-personas}, which are behaviour patterns defined by correlations among multiple variables in the data \citep{Goodwin2009}. The proto-personas are user models that serve as the foundation of the actual personas, which are produced by stepping through the remaining stages of Goodwin's method. To examine the performance of the proposed method, proto-personas are produced manually and by the DOC algorithm. In an empirical comparison, we attempt to demonstrate that similar results are obtained with the manual and mixed method, but that the DOC algorithm is more rigorous and will find optimal clusters. Afterwards, we attempt to demonstrate that the information contained in these clusters is similar to the relationships displayed by perceptual maps produced by running MCA on a category-binned version of the same data set. Lastly, we discuss approaches to handling similar subspace clusters as well as the process of turning the proto-personas into final personas and strength and limitations of the method. The LSA method was not included in this study due to multiple reasons. The data set used did not contain a strict structure; thus, a labour-intensive process was required to manually link text segments to the question they answer. Besides, the data set contained missing values, which the LSA method in the version described by \cite{Miaskiewicz2008} is currently unable to handle. We also had problems locating general semantic spaces in the language (Danish) that the interviews were conducted in.

%HIGH DIM PROBLEM

%However, it is required that a semi-automatic method should produce similar results to a human designer when finding groups with a similar size to ensure the validity of this approach. Thus, in the present study, we investigated the development of proto-personas using Goodwin's method \citep{Goodwin2009, Cooper2007, Goodwin2002} by performing the segmentation stage twice, where we used a manual approach and a semi-automatic quantitative segmentation technique.

\section{Methods}

\subsection{Data}

The data used for persona development comprised of semi-structured interviews from a sample of older adults from 20 households living under different circumstances throughout Denmark. The older adults were self-recruited through community centres and leisure activities for older people, as well as via an advertisement in a magazine for older people. Interviewees were scheduled with the older adults who responded based on the following pre-specified characteristics: residence (must be living in own home), gender (a ratio of 50:50 was the aim), age (65 years and above), geography (urban and non-urban representatives), household inhabitants (co-living or single living), and food service (including meals-on-wheels subscribers and non-subscribers). Older adults with major cognitive impairments or physical disabilities were not included in the study. Among the 20 interviews, two were conducted as family interviews where both the husband and wife participated, whereas the other 18 were individual interviews. In both family interviews, one of the participants dominated the conversation, and the dominating participant was thus selected as the representative of the household.

Using narrative-inspired interview techniques \citep{Jovchelovitch2000}, the older adults were asked about their meal experiences, meal preferences, everyday life, skills, desires, and opinions of specific intervention suggestions in their own homes. The interview also included questions that the participants could answer using a numeric rating scale. For example, the participants were asked, ``How important is the price of food to you on a scale from 0 to 10, where `it holds no importance' is 0 on the scale and `it has significant importance' is 10?'' After the participant rated the question, the interviewer followed up on the question by asking why that particular rating was given. The interviews were recorded and then transcribed with an id (1--20) before further analysis.

\subsection{Manual personas}

A PD with a background in food science and anthropology was hired to perform proto-personas manually and did so by performing stage one to three of Goodwin's method \citep{Cooper2007, Goodwin2002}. The PD was not involved with conducting or transcribing the interviews.

\begin{figure}[!h]
\includegraphics[angle=0,scale=3.0]{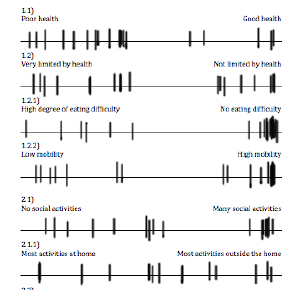}
\caption{Subset of the VASs with all markings visible.}
\label{fig:scales}
\end{figure}

Behavioural variables were identified (stage one) by content analysis \citep{Bjrner2015}. The interviews were coded by the PD using the Atlas.ti software. The PD labelled attitudes, aptitudes, motivations, skills, and activities according to Goodwin's method \citep{Cooper2007}. A list was compiled of the most commonly recurring coding labels across the interviews, which served as behavioural variables. After removing redundant or irrelevant variables, the final number of dimensions was 47 (numbered as d1 to d47). The interviewees were mapped to the behavioural variables (stage two) using a VAS for each variable. Goodwin suggested labelling each interviewee with a letter \citep{Goodwin2009}, but this was impractical because the overview was lost due to clutter when placing all interviewees on a single VAS. Thus, Adobe Photoshop CS6 was used instead to visualize the scales, which features a layering system that works similarly to a stack of transparent sheets. Each layer could be drawn and made visible or invisible. The scales were listed below each other on the background layer as black horizontal lines. Each end of the line represented an extreme (e.g., never cooks the main meal at one end and cooks the main meal daily at the opposite end), which was written in text beside the line. The behaviour of each interviewee was then marked on a unique layer by a black vertical line on each scale (Figure~\ref{fig:scales}). The placement of the line was based on the PD's judgment after another reading of the coded interview. All of the layers were saved in separate image files for subsequent analysis using a software algorithm.

% Figure

The process employed for identifying significant behaviour patterns (stage three) involved systematically comparing the markings on the layers. Each layer contained the markings on all (or nearly all) of the behavioural variable scales for one interviewee. The behaviour of each interviewee could then be systematically compared with that of others by controlling the visibility of the layers. If the markings of two interviewees were located close to each other on a scale, the behaviour of the interviewees was considered more similar compared with those interviewees with more distant marks. If interviewees had more markings close to each other on several scales, they were also considered more identical than interviewees with a few similar VAS markings. Without any restrictions on group size, the PD employed the following systematic approach during manual segmentation.
\begin{enumerate}
\item All interviewee layers were made visible, and a note was made of the scales that seemed most ``polarized'' in terms of the most markings at the extremes.
\item One of the polarized scales was selected, and all of the interviewees were initially assigned to one of two groups characterized by the extremes of the scale.
\item The interviewee layers for the members in one of the extreme groups were then made visible, and the other scales, where the majority of the group had markings in close proximity, were noted, whereas the members who ``stood out'' on the scales were deleted.
\item Step 3 was repeated for the other extreme group.
\item Steps 2--4 were repeated until each polarized scale was used to represent at least one group.
\item The groups obtained were checked for redundancy.
\end{enumerate}

Another human PD (henceforth denoted PD2) with a background in engineering was hired to redo the segmentation step manually to account for human bias. Similar to the first PD (henceforth denoted PD1), PD2 studied the interview recordings, transcripts, and the scales and was instructed to form groups without any restrictions on group size. PD2 described how she applied the following method for segmenting the interviewees manually.
\begin{enumerate}
\item The layer with markings of the two first interviewees (interviewee 1 and 2) were made visible, and depending on whether the markings on the scales of the two visible layers appeared in close proximity the second layer would remain visible or hidden away. The layer of the next interviewee (interviewee 3) would then be displayed and compared to the visible layer(s) and depending on similarity kept on or turned off. This approach was then continued with the rest of the interviewees, where those who did not fit in were turned off again and those who had many of the same answers on the scales remained on.
\item Once all comparisons had been made and a group had been formed, the process was repeated with the first interviewee that was omitted from the group, which was then compared to all other interviewees to form another group.
\item The few participants that remained without a group were compared to find scales with common markings. Those with most markings in common where grouped and used as a starting point for layer-comparisons with all other interviewees one by one to grow the group.
\item When all interviewees were in a group, the interview recordings were revisited to confirm the segmentation.
\end{enumerate}

\subsection{DOC-based personas}

%A small custom software program (internally named scales-to-digits) was written, which looked over the pixels of each of the interviewee layer image files to identify the markings. The program identified each scale, determined if a marking had been placed, and then extracted the position of it relative to the scales length. Checkboxes would also be identified and the pixel values inside of it would determine whether it had been checked or not.

The clustering process employed the VASs produced from stage one to stage three for the manual personas. The relative positions along the VASs were converted into scores between 0 and 1 using custom image-processing software. The extracted data were then analysed by the DOC algorithm \citep{Procopiuc2002}. The open-source implementation of the DOC algorithm found in the ELKI Data Mining Framework \citep{Achtert2008} was altered slightly to achieve overlapping (e.g., one interviewee could appear in more than one cluster), while still ensuring full coverage (all interviewees must appear in a cluster). Details of the changes made to the DOC algorithm are described in the following.

% Figure
\begin{figure}
\includegraphics[angle=0,scale=0.34]{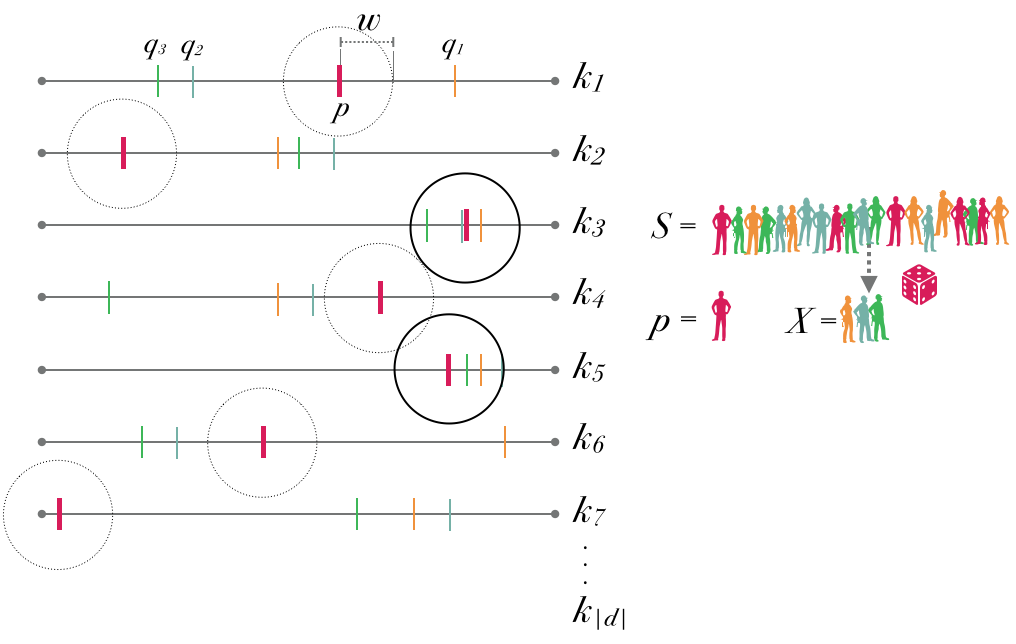}
\caption{Visual example of the entire data set ($S$), a random sample ($p$), and the discrimination set ($X$) in this case consisting of the random samples $q_1,q_2,q_3$. From the set of behavioral dimensions ($k_1$ to $k_{|d|}$), $p$ and the members of $X$ cluster in a subset/subspace consisting of the dimensions $k_3$ and $k_5$.}
\label{fig:docdesc1}
\end{figure}

\begin{figure}
\includegraphics[angle=0,scale=0.34]{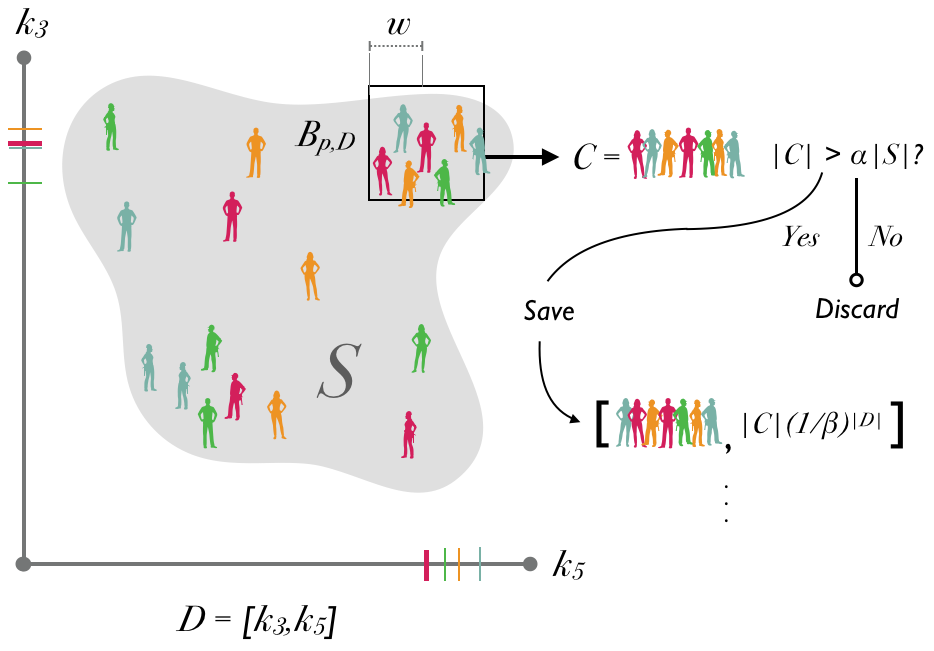}
\caption{Visual example of the two-dimensional cluster subspace ($D$). A hyper-rectangle ($B_{p,D}$) with edges of length $2w$ is defined around $p$. Every member of $S$ is hit-tested against $B_{p,D}$ to produce a subspace cluster $(C,D)$. If the amount of members in $(C,D)$ exceeds a threshold ($\alpha |S|$) the subspace cluster is saved and a quality score is computed ($|C|(1/\beta)^{|D|}$).}
\label{fig:docdesc2}
\end{figure}

DOC uses random sampling to obtain approximated optimal clusters \citep{Procopiuc2002}. Let the entire data set be denoted by $S$. The DOC algorithm starts with a randomly selected sample from $S$, which we denote as $p$. A small subset of data points from $S$ is then obtained by random sampling, which is called the discrimination set denoted by $X$. The algorithm uses $X$ to determine the subspace of a cluster to which $p$ belongs by iterating over all dimensions $d$ and keeping each dimension where the data points in $X$ are within a distance $w$ from $p$ (see Fig.~\ref{fig:docdesc1}). This is also expressed by the formula: $| q_k - p_k | \leq w, \forall q \in X, \forall k \in d$. $w$ is half the maximum length of the data points in a cluster that can be displaced from each other and it must be provided as a parameter for the algorithm. After identifying the dimensions comprising the cluster subspace, all data points in the data set are tested for cluster membership by investigating whether they are within the hyper-rectangle (denoted $B_{p,D}$) around $p$ of length $2w$. If the fraction of data points in $B_{p,D}$ exceeds a density threshold (denoted by $\alpha$), then the clustered data points $C$ and the subspace $D$ are saved (see Fig.~\ref{fig:docdesc2}). This is the core of the DOC algorithm, which is then repeated iteratively to determine an optimal cluster in the data set. An optimal cluster is defined by the ratio between the number of data points in the cluster $|C|$ and the number of dimensions of the subspace where the cluster exists $|D|$. The parameter $\beta$ is used to calculate a quality score for each cluster with the following formula: $|C|(1/\beta)^{|D|}$. The DOC algorithm returns the cluster with the maximum quality score.

% Figure

The ELKI implementation of DOC does not handle overlapping, although the DOC algorithm is capable of handling it \citep{Procopiuc2002}. By default, ELKI finds the optimal clusters among the set of unclustered interviewees until all of the interviewees have been clustered. To enable overlapping, we followed the directions given by the authors of the DOC algorithm and kept clustered interviewees in the set, which required a new end condition. We used an end condition that ensured full coverage (all interviewees were in a cluster). The aim of DOC involves finding one optimal cluster, which requires running two nested loops comprising an outer loop and an inner loop. An interviewee is picked at random in the outer loop and evaluated against several randomly generated discriminant sets in the inner loop (sets of interviewees) to determine the relevant subspaces. Clusters including the selected interviewee are then generated in the subspaces identified. The most optimal cluster among all the generated clusters is finally returned. To ensure full coverage, our change aimed to select the same interviewee in the outer loop instead of random selection. Therefore, the algorithm required another parameter $o$ as the data point around which an optimal cluster was found. Running DOC with the change made it more goal-oriented as it would find the optimal cluster containing a specified interviewee rather than an arbitrary optimal cluster. DOC was then run for each interviewee until an optimal cluster was found for each of them. The modified DOC algorithm is presented as pseudo-code in ALGORITHM~\ref{alg:one}.

% Algorithm
\IncMargin{1em}
\begin{algorithm}[t]
\DontPrintSemicolon
\KwIn{Set of subjects $S$, set of dimensions $d$, max distance $w$, density threshold $\alpha$, and parameter $\beta$ used to define optimal clusters.}
\KwOut{Set of non-redundant optimal clusters formed around each subject.}
$r = | \log (2|d|) / \log (1/2\beta) |$\;
$m = (2/\alpha )^r\ln (4)$\;
\SetKwFunction{FDoc}{DOC}
\Begin{
\ForEach{\upshape subject $o$ \upshape in $S$}{
$(C_o,D_o)$ = \FDoc{$S$, $w$, $\alpha$, $\beta$, $o$}\;
}
\KwRet non-redundant set of all optimal clusters $(C_o,D_o)$\;
}
\;
\SetKwProg{Fn}{function}{:}{}
\Fn{\FDoc{$S$, $w$, $\alpha$, $\beta$, $o$}}{
\For{$i=1$ \KwTo $2/\alpha$}{
Set $p = o$ /* instead of random sampling */\;
\For{$j=1$ \KwTo $m$}{
Choose $X \subseteq S$ with size $r$ uniformly at random\;
$D = \{k|| q_k - p_k | \leq w, \forall q \in X, \forall k \in d\}$\;
$C = S \cap B_{p,D}$\;
\If{($|C| < \alpha |S|$)}{$(C,D) = (\emptyset ,\emptyset )$\;}
}
}
\KwRet cluster that maximizes $|C|(1/\beta)^{|D|}$ over all computed clusters $(C,D)$\;
}
\caption{Modified DOC algorithm based on that proposed by \cite{Procopiuc2002}}
\label{alg:one}
\end{algorithm}
\DecMargin{1em}

The adjusted DOC algorithm was used with parameter settings of $\alpha = 0.1$ (at least two points in clusters when the sample size was 20), $w = 0.3$ (just below $1/3$ of the length of each scale), and $\beta$ was 0.25, 0.45, 0.65, and 0.85 in different runs. The $\beta$ parameter had a direct influence on the size of the discrimination set used to identify the subspace. When it was set to 0.25, 0.45, 0.65, and 0.85, the algorithm used two, three, four, and five interviewees, respectively, to select the subspace (see lines 1 and 14 in ALGORITHM~\ref{alg:one}). %The high-member priority runs ($\beta$ = 0.65 and 0.85) demonstrated how the DOC algorithm could find clusters with many participants in subspaces that shared as many dimensions as possible, whereas the high-dimension priority runs ($\beta$ = 0.25 and 0.45) produced clusters with a size that was more comparable to those selected by the PD.

\subsection{MCA-based personas}
Laporte et al.'s method \citep{Laporte2012} was used to generate a perceptual map displaying the interviewees in terms of how they vary the most. Originally, Laporte et al. used a strict questioning structure to collect open-ended textual responses to a fixed set of questions. The resulting answers to each question were afterwards manually converted into categories. Since the core contribution of the work by Laporte et al. is the proposal of using MCA to guide the segmentation, a different approach was used in this study to arrive at the categorical data.

The scores from the VASs were converted into categories by dividing the scales into two or three bins and then assigning each marking falling within a bin a category \citep{Abdi2007}. Two bins were used when the VAS tended to be polarized where all the markings were located towards either end, while three bins were used when markings were more uniformly distributed along the VAS. As an example, the VAS representing the degree to which the older adult engaged in activities outside or inside the home was divided into three bins resulting in markings receiving the category $StaysAtHome$, $BothAtHomeAndOutside$, or $Outgoing$. A resulting table had rows representing each interviewee and each column representing a behavioural variable as a category instead of a number between 0 and 1. The table was used as an input to the MCA function in the FactoMineR package for R \citep{Husson}. The results of the MCA were compared with the subspace clusters by inspection of the produced perceptual maps. Further insights were gained by investigating correlations between categorical variables and the two dimensions making up the perceptual map \citep{Laporte2012}. As MCA handles missing data by creating separate categories for the missing data, the perceptual map will also be influenced by how similarly the interviewees omit data. %Experimentation was done with the missMDA package for R \citep{Husson2018}, which contains functionality to impute the missing data based on non-missing entries. However, the results from the original MCA turned out to be more similar to those of the manual grouping and were therefore used for the comparison.

\section{Results}

\begin{table}[!b]%
\caption{PD1 groups}
\label{tab:pdgroups}
\begin{minipage}{\columnwidth}
\begin{center}
\begin{tabular}{ccp{0.5\linewidth}}
Group ID & Member IDs & Short pattern description\\\toprule
PD1A & 6,16,20,(8) & Social older adult who enjoys and engages in social meals, living with a spouse.\\
PD1B & 1,12,17 & Single female: Poor health, mostly at home, depending on others, enjoys social meal, eats at elder caf\'e but always eats alone.\\
PD1C & 4,7,(3),(15) & Single female, with eating difficulties, low sensory input, always eats alone, and receives meals-on-wheels.\\
PD1D & 11,13,14 & Single female, very active, eats a lot at elder caf\'e, and enjoys life.\\
PD1E & 10,18,(5) & Single male, very active and social, likes IT, always eats at elder caf\'e, and enjoys life.\\
PD1F & 2,6,18,(9) & Enjoys cooking meals and is positive about the idea of delivered groceries.\\
PD1G & 16,19 & Living with a spouse, money is not an issue, and values food quality and ecology.\\
\bottomrule
\end{tabular}
\end{center}
\bigskip\centering
\footnotesize
\emph{Note:} A member ID in parenthesis indicate that the interviewee had similar markings in terms of most of the common scales but not all of them.
\end{minipage}
\end{table}%

\begin{table}[]%
\caption{PD2 groups}
\label{tab:pdgroups2}
\begin{minipage}{\columnwidth}
\begin{center}
\begin{tabular}{ccp{0.5\linewidth}}
Group ID & Member IDs & Short pattern description\\\toprule
PD2A & 2,5,6,8,9,16,20 & This group is generally in good health and independent. They come out a lot and are active both socially and physically. They often eat with others and can cook and shop for themselves. They enjoy life and are not limited by their age in everyday life.\\
PD2B & 1,3,10,17,(12),(14),(15) & This group is plagued by ill health and is physically restricted. They do not enjoy food and see it as a means of survival. They are inactive and often receive outside help. Also, they do not have much appetite and only sometimes make the food themselves. They are often alone when they eat, but enjoy when there is something social - however, they do not have the largest network and they do not come out as much.\\
PD2C & 4,7,14,(12) & This group receives food from the outside and rarely makes the food themselves. For them, food is a means of survival and they often eat alone. They are most at home and dependent on others in everyday life. Also, they have few closest contacts.\\
PD2D & 11,13,19,(18) & This group loves food and has a good appetite and cares about the quality of the food. They can mostly manage on their own, but still feel limited by their health in everyday life. They like to get out and be social and generally enjoy life.\\
\bottomrule
\end{tabular}
\end{center}
\bigskip\centering
\footnotesize
\emph{Note:} A member ID in parenthesis indicate that the interviewee had similar markings in terms of most of the common scales but not all of them.
\end{minipage}
\end{table}

The final groups produced manually by the PDs are listed with the group ID and members in Table~\ref{tab:pdgroups} and Table~\ref{tab:pdgroups2}. The group members in parentheses were placed there by the PD to indicate that the interviewee had similar markings in terms of most of the common scales but not all of them. As can be seen from the groups found through the manual process, they are rather different. PD1 has found groups that contain a maximum of three interviewees per group (excluding the interviewees in parentheses), while PD2 has aimed for fewer and larger groups. 

%\subsection{High dimensional priority subspace clustering}

DOC is based on random sampling, so we started by running the DOC algorithm several times and concluded that we observed no noticeable variations in the resulting clusters. The 61 clusters obtained after running the subspace-clustering algorithm with the selected parameters are shown in Table~\ref{tab:smallsubspace} ($\beta=0.25$), Table~\ref{tab:large45subspace} ($\beta=0.45$), Table~\ref{tab:large65subspace} ($\beta=0.65$), and Table~\ref{tab:large85subspace} ($\beta=0.85$). The clusters are shown in the tables, where the columns represent each cluster and the rows are the behavioural variables. Each cell displays either the cluster mean on the particular dimension or NA if the dimension is not a part of the subspace where the cluster was found. %, In addition, the proposed similarity score (Equation~\ref{eq:similarity_score}) was applied with hierarchical clustering to obtain a dendrogram (Figure~\ref{fig:simall}) providing an overview of the relationship between subspace clusters.

An optimal cluster was generated for each interviewee per run, which produced more clusters than those found by each of the PDs. After searching for the groups determined by the PDs in the set of subspace clusters, we obtained the comparison shown in Table~\ref{tab:pdvssubspace}.

\begin{table}[]%
\caption{PD groups compared with subspace clusters}
\label{tab:pdvssubspace}
\begin{minipage}{\columnwidth}
\begin{center}
\begin{tabular}{cc|cc}
\multicolumn{2}{c}{PD manual groups} & \multicolumn{2}{c}{Closest subspace cluster}\\
Group ID & Member IDs & Cluster ID & Member IDs\\\toprule
PD1A & 6,16,20,(8) & A45 & 6,16,20\\
PD1B & 1,12,17 & J45 & 1,12,17\\
PD1C & 4,7,(3),(15) & O25 & 4,7\\
PD1D & 11,13,14 & N45 & 11,13,14\\
PD1E & 10,18,(5) & P25 & 10,18\\
PD1F & 2,6,18,(9) & - & -\\
PD1G & 16,19 & - & -\\\midrule
PD2A & 2,5,6,8,9,16,20 & A85 & 6,8,9,16,20\\
 & & B85 & 2,6,8,16,20\\
 & & F85 & 2,5,6,16,20\\
PD2B & 1,3,10,17,(12),(14),(15) & M65 & 1,3,12,17\\
PD2C & 4,7,14,(12) & O25 & 4,7\\
PD2D & 11,13,19,(18) & H25 & 11,13\\
\bottomrule
\end{tabular}
\end{center}
\end{minipage}
\end{table}%

\begin{figure}
    \centering
    \begin{minipage}[t]{0.47\textwidth}
        \centering
        \includegraphics[width=1.0\textwidth]{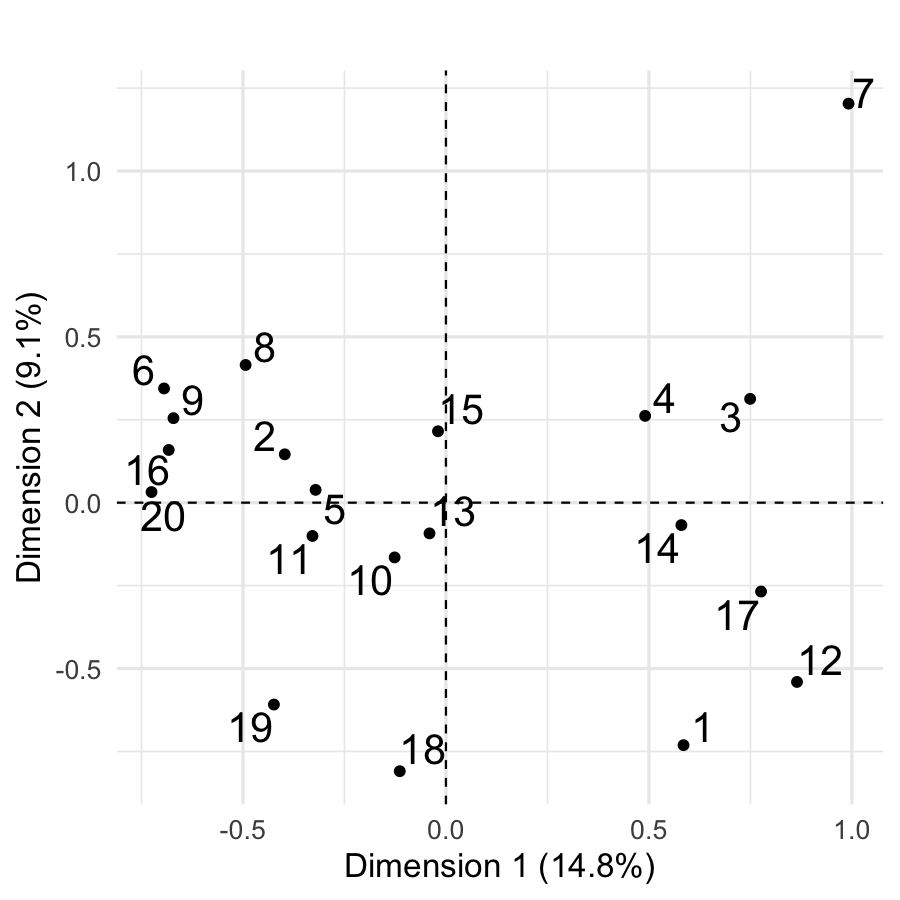} % first figure itself
        \caption{MCA perceptual map produced by the MCA over all categorical variables.}\label{fig:MCA}
    \end{minipage}\hfill
    \begin{minipage}[t]{0.47\textwidth}
        \centering
        \includegraphics[width=1.0\textwidth]{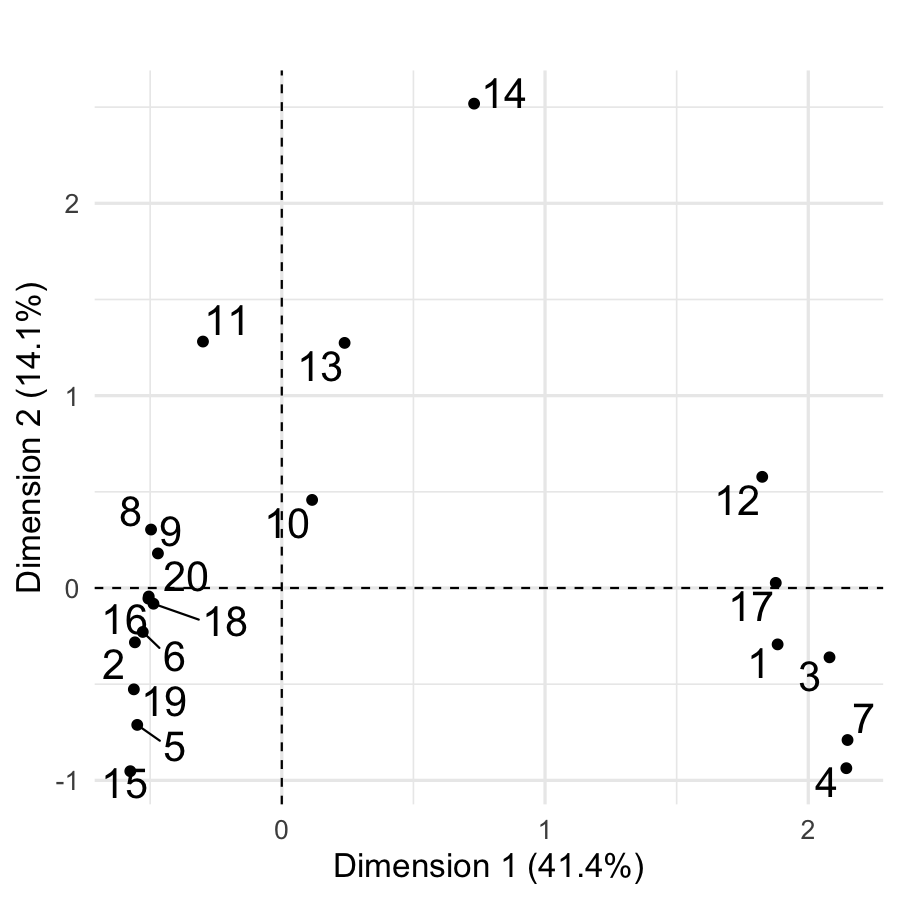} % second figure itself
        \caption{Contingency perceptual map produced by a correspondence analysis of the contingency table on how often two interviewees appear in the same subspace cluster across all the subspace clusters except J85, K85, L85, and M85.}\label{fig:CA}
    \end{minipage}
\end{figure}

The MCA was run on the binned data and the first two dimensions explained 14.8\% and 9.1\% of the variance in the data. The dimensions make up a perceptual map where interviewee relationships are visualized as distances. As displaying the subspace clusters directly on the perceptual map of the MCA would result in a packed and unreadable visualization, the comparison was done by investigating how related interviewees were by how often they co-occur in a subspace cluster. A contingency table (Table~\ref{tab:membercounts}) was created with counts of member co-occurrences across all subspace clusters found (except J85, K85, L85, which are the result of trying to find too large non-existing clusters). A correspondence analysis was performed on the contingency table to obtain a perceptual map of the interviewee relationships (Figure~\ref{fig:CA}) and allow comparison with the perceptual map produced by the MCA (Figure~\ref{fig:MCA}). Both perceptual maps were interpreted in the same manner: Interviewees in close proximity to each other are more related.

\begin{table}[]
\centering\scriptsize
\setlength\tabcolsep{2pt} % General space between columns (6pt standard)
\renewcommand{\arraystretch}{1.0} % General space between rows (1 standard)
\caption{Contingency table with counts of how often two interviewees co-occur in a subspace cluster over all subspace clusters except J85, K85, L85, and M85.}
\label{tab:membercounts}%

\begin{tabular}{l|llllllllllllllllllll}
   & 1  & 2 & 3 & 4 & 5  & 6  & 7 & 8 & 9  & 10 & 11 & 12 & 13 & 14 & 15 & 16 & 17 & 18 & 19 & 20 \\\hline
1  & 10 & 0 & 3 & 4 & 0  & 0  & 1 & 0 & 0  & 2  & 0  & 2  & 1  & 0  & 0  & 0  & 8  & 0  & 0  & 0  \\
2  & 0  & 5 & 0 & 0 & 1  & 3  & 0 & 1 & 0  & 0  & 0  & 0  & 0  & 0  & 0  & 4  & 0  & 0  & 0  & 5  \\
3  & 3  & 0 & 4 & 1 & 0  & 0  & 0 & 0 & 0  & 0  & 0  & 1  & 0  & 0  & 0  & 0  & 4  & 0  & 0  & 0  \\
4  & 4  & 0 & 1 & 5 & 0  & 0  & 2 & 0 & 0  & 0  & 0  & 0  & 0  & 0  & 0  & 0  & 3  & 0  & 0  & 0  \\
5  & 0  & 1 & 0 & 0 & 11 & 4  & 0 & 0 & 0  & 1  & 0  & 0  & 0  & 0  & 4  & 6  & 0  & 0  & 2  & 7  \\
6  & 0  & 3 & 0 & 0 & 4  & 18 & 0 & 2 & 6  & 1  & 0  & 0  & 1  & 0  & 2  & 18 & 0  & 3  & 2  & 13 \\
7  & 1  & 0 & 0 & 2 & 0  & 0  & 3 & 0 & 0  & 0  & 0  & 1  & 0  & 0  & 0  & 0  & 2  & 0  & 0  & 0  \\
8  & 0  & 1 & 0 & 0 & 0  & 2  & 0 & 8 & 4  & 0  & 2  & 0  & 1  & 0  & 0  & 7  & 0  & 0  & 0  & 8  \\
9  & 0  & 0 & 0 & 0 & 0  & 6  & 0 & 4 & 12 & 2  & 1  & 0  & 2  & 0  & 0  & 12 & 0  & 1  & 1  & 10 \\
10 & 2  & 0 & 0 & 0 & 1  & 1  & 0 & 0 & 2  & 7  & 0  & 0  & 1  & 1  & 0  & 2  & 1  & 1  & 0  & 3  \\
11 & 0  & 0 & 0 & 0 & 0  & 0  & 0 & 2 & 1  & 0  & 5  & 0  & 2  & 1  & 0  & 3  & 0  & 0  & 0  & 3  \\
12 & 2  & 0 & 1 & 0 & 0  & 0  & 1 & 0 & 0  & 0  & 0  & 5  & 1  & 1  & 0  & 0  & 5  & 0  & 0  & 0  \\
13 & 1  & 0 & 0 & 0 & 0  & 1  & 0 & 1 & 2  & 1  & 2  & 1  & 7  & 2  & 0  & 3  & 2  & 0  & 0  & 2  \\
14 & 0  & 0 & 0 & 0 & 0  & 0  & 0 & 0 & 0  & 1  & 1  & 1  & 2  & 3  & 0  & 0  & 1  & 0  & 0  & 0  \\
15 & 0  & 0 & 0 & 0 & 4  & 2  & 0 & 0 & 0  & 0  & 0  & 0  & 0  & 0  & 4  & 3  & 0  & 0  & 0  & 1  \\
16 & 0  & 4 & 0 & 0 & 6  & 18 & 0 & 7 & 12 & 2  & 3  & 0  & 3  & 0  & 3  & 31 & 0  & 3  & 2  & 24 \\
17 & 8  & 0 & 4 & 3 & 0  & 0  & 2 & 0 & 0  & 1  & 0  & 5  & 2  & 1  & 0  & 0  & 12 & 0  & 0  & 0  \\
18 & 0  & 0 & 0 & 0 & 0  & 3  & 0 & 0 & 1  & 1  & 0  & 0  & 0  & 0  & 0  & 3  & 0  & 4  & 0  & 2  \\
19 & 0  & 0 & 0 & 0 & 2  & 2  & 0 & 0 & 1  & 0  & 0  & 0  & 0  & 0  & 0  & 2  & 0  & 0  & 4  & 3  \\
20 & 0  & 5 & 0 & 0 & 7  & 13 & 0 & 8 & 10 & 3  & 3  & 0  & 2  & 0  & 1  & 24 & 0  & 2  & 3  & 29
\end{tabular}
\end{table}

\section{Discussion}

\subsection{Results comparison}

The two PDs agreed on a set of cluster memberships. PD1A and PD2A both shared four members (6, 16, 20, 8), and the descriptions point to a group of older adults that engages in social meals and enjoys life. PD1B and PD2B shared three members (1, 12, 17) and both PDs described the group as dependent on others, enjoying the company of others but usually eating alone. PD1C and PD2C shared two members (4, 7) and the descriptions both captured that the groups receive meals-on-wheels. PD1D and PD2D also shared two members (11, 13), but the descriptions did not clearly show common features. In general, the number of groups and amount of members deviated to a high degree when the segmentation process was performed manually. The strategy applied by each PD to segment the interviewees was also different and not always optimal. For instance, PD2 used a chronological approach to compare the interviewees, which made the VAS markings of the initial interviewee selected for comparison more influential on the final cluster. PD1 attempted to avoid this bias, by an initial step in which she selected dimensions with a high concentration of markings at the extremes. However, focusing on the extremes fails to capture clusters around the centre of the VAS.

Subspace clustering obtained five of the seven groups identified manually by PD1 (not counting the partial members). The subspaces in which the five clusters were found also matched the short descriptions given by PD1 but add more details. As an example, PD1 described group PD1D comprised of interviewees 11, 13, and 14 as, ``Single female, very active, eats a lot at elder cafe and enjoys life.'' A short description of subspace cluster N45 (in Table~\ref{tab:large45subspace}) consisting of the same members could be written as, ``Single (d46) female (d47) with no interest in IT (d12), who has high mobility (d4) with many activities outside the home (d6) and who exercises regularly (d41). She is often in contact with family members (d7), eats three meals a day (d37) sometimes alone but other times with others (d14) and she prefers to eat with others but is also happy with not always having company for the meal (d13). She often eats out (d15) but finds it appealing to get a box delivered with groceries for cooking the main meal (d24).'' Thus, the reader of these descriptions gets the impression of the same kind of person, but some details are added in the case of the subspace clustering version.

The other two clusters were not found by the algorithm because the number of dimensions, where the markings of the members were below a distance of $w$ (0.3) was less compared to other member constellations. For example, the cluster comprising interviewees 2, 6, and 18 (PD1F) had markings with a distance below $w$ for 10 dimensions. The subspace clusters found by the algorithm (with $\beta=0.45$) containing either interviewees 2, 6, or 18 had a larger number of dimensions when combined with the other interviewees, i.e., E45, A45, and K45 had 18, 24, and 14 dimensions, respectively. Thus, E45, A45, and K45 were of better quality (more optimal) than the potential cluster comprising 2, 6, and 18, which was ignored when one of the other clusters was found by the algorithm. PD1 described group PD1F as ``Enjoys cooking meals and is positive about the idea of delivered groceries.'' The dimension related to cooking enthusiasm only occurred in a few of the subspace clusters (J25 and P25), and it was related to moderate enthusiasm (around 0.6). The reason for selecting this group was not related to the number of dimensions where the grouping was found but instead to the specific dimensions. Interestingly, the PD found this group because it highlighted a set of interviewees with extreme markings on a particular dimension related to the purpose of personas. If some variables are more important than others, then the DOC algorithm will no longer be helpful because it is designed to find the most optimal cluster in terms of the number of dimensions and members as controlled by the parameter $\beta$.

For all the four groups formed by PD2, similar subspace clusters were found by the DOC algorithm which all missed a single of the members PD2 suggested (when not counting the partial members). The PD2A cluster appears to be aligned with both the subspace cluster results (Figure~\ref{fig:MCA}) and the results from the MCA (Figure~\ref{fig:MCA}). However, the chronological method used by PD2 to segment the interviewees entailed exclusion of members from groups formed early, into groups formed late in the process. Thus, core members of PD2A and PD2B do not appear in PD2C or PD2D, which become groups formed from ungrouped interviewees.

Only ten of the 61 subspace clusters obtained by the DOC algorithm matched groups found by the PDs. This low hit rate can be explained by the multiple runs with a different $\beta$ parameter, but also by the PDs tendency to relax the criteria for determining whether an interviewee was a member of a group. Interviewees surrounded by parentheses in the tables were determined by the PDs as being similar to the core members of a group for the majority (but not all) of the dimensions in which the core members had clustered markings. Thus, fewer groups were required by the PDs to achieve total coverage, while the subspace clustering produced one cluster per interviewee (per $\beta$ run) unless the members and subspace of the clusters were completely identical.

The DOC algorithm produced a set of less useful subspace clusters (J85, K85, L85, and M85) (Table~\ref{tab:large85subspace}). Each of these clusters contained over half of the participants, but, similar to the “population of Europe” example given in the introduction, the subspace of the clusters comprised a maximum of three behavioural dimensions. Omitting these trash clusters would leave participants 1, 3, 4, 7, 12, 14, and 17 outside all of the clusters. This is because the high $\beta$ parameter, which was set to 0.85, yielded subspace clusters containing five or more participants. If every participant must be clustered with five or more participants, this will produce groups that share very few dimensions.

Finally, the DOC algorithm obtained unique clusters that were not found by the PDs. For example, the PD did not identify any proto-personas comprising interviewees 5 and 15, such as subspace cluster L25. The L25 cluster describes a companionless (d46) older man (d47) with some health problems (d1) but who still has high self-sustainability (d2, d3, d4, d20, d22, d23, and d42) and good understanding (d10) and interest (d12) in IT technology. He has a relatively large social network (d9) but balances his time between social activities (d5) and alone time at home (d6). In particular, he prefers to cook (d29) in his kitchen (d31) and to eat alone (d13, d14, and d19). None of the groups found by the PD included the trait of preferring to eat alone.

\begin{figure}
\includegraphics[angle=0,scale=0.25]{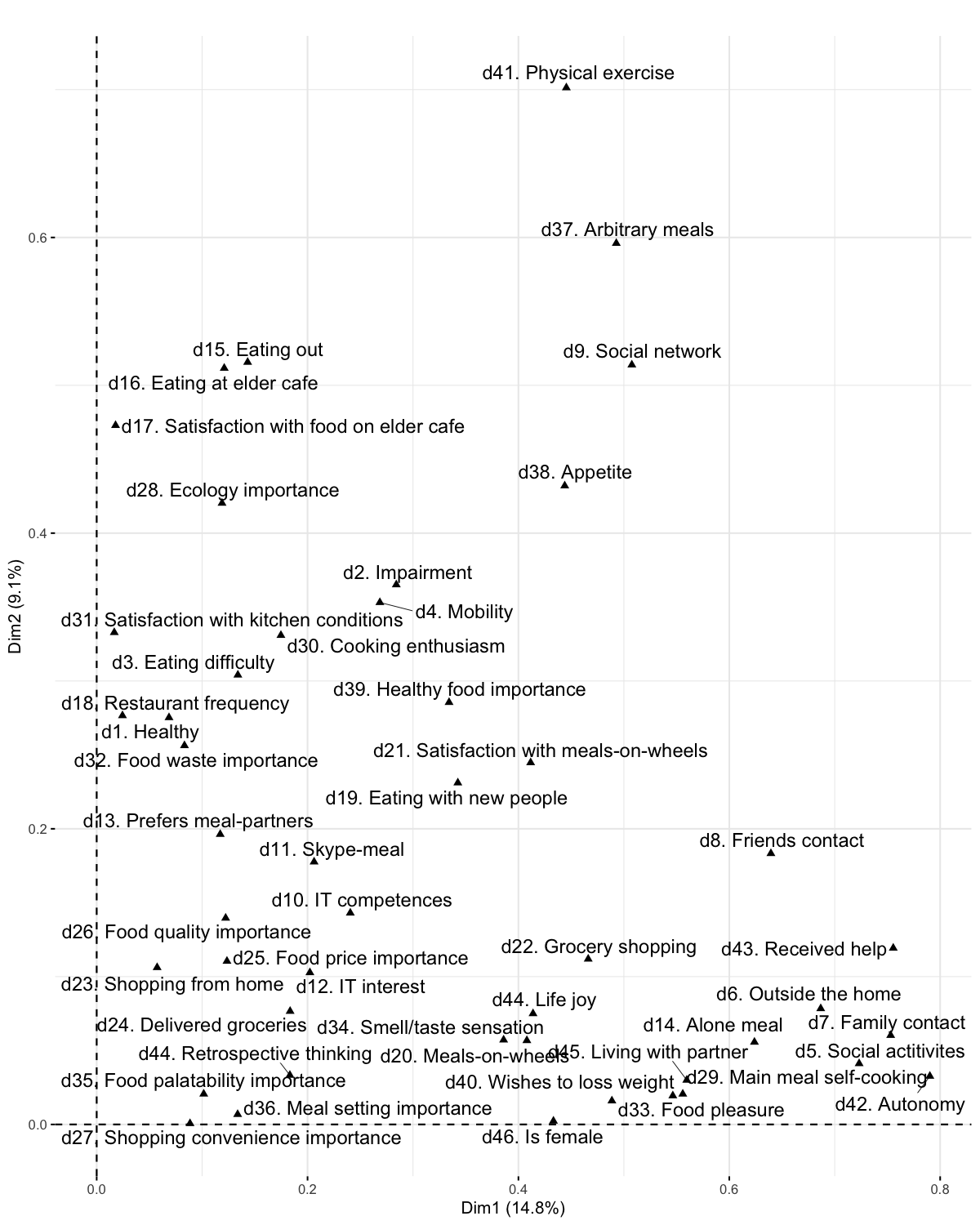}
\caption{Plot displaying the degree of correlation between each categorical variable and the first two dimensions produced by the MCA.}
\label{fig:correlation}
\end{figure}

In general, the contingency map from the subspace clusters shows a more caricatured relationship between interviewees compared to the MCA map due to the low dimensional data used to generate it. However, similar relationships can be seen. The x-axis separates the two main groups (interviewees 1, 3, 4, 7, 12, 14, and 17 from interviewees 2, 5, 6, 8, 9, 11, 16, and 20) in both maps. Investigating how correlated each categorical behavioural variable is with the two axes in the MCA plot yielded Figure~\ref{fig:correlation}. The first axis (and thus the most important points of distinction) is highly correlated with categorical variables representing degrees of autonomy behaviour displayed by each interviewee (e.g., amount of received help, self-cooking, and social activities). Thus, we see that subspace clustering makes clusters that account for the largest variances in the data. However, the variance on the second axis of the MCA plot does not appear in the contingency map. According to Figure~\ref{fig:correlation}, the second axis tends to capture the tendency to eat out, tendency to eat at elder caf\'e and satisfaction with the food on the caf\'e. Level of physical exercise and the tendency of arbitrary meals have a high correlation with the second axis, but also factor into the first axis.

% Figure

\subsection{Hyperparameter selection}

None of the PDs specified at what distance they considered two interviewees to be similar/dissimilar, but instead relied on how human perception and the natural ability to perceive objects as grouped (Gestalt law of grouping). The $w$ parameter represents the maximum distance two interviewees can be apart before they are classified as dissimilar. \cite{Procopiuc2002} proposed a method for deriving $w$ by first calculating the average distance between an interviewee and the nearest neighbour across dimensions ($ w_i = \sum_{j=1}^{|d|}|q_{j}^{i}-p_{j}^{i}|/|d|$) and then taking the average of the resulting averages across interviewees ($ w = \sum_{i=1}^{|n|}|q_{j}^{i}-p_{j}^{i}|/|d|$). Applying the same calculations on our dataset resulted in a $w$ of 0.2826185 which was rounded up to 0.3 for simplicity.

To gain full coverage, the PDs were instructed that every interviewee should be part of a group. Similarly, the density threshold represented by the $\alpha$ parameter was set to 0.1 at all runs, which entailed accepting a cluster if it contained at least two members (at a sample size of twenty interviewees). The $\alpha$ parameter should be adjusted ($\alpha = |C_{min}|/|S|$ where $|C_{min}| \leq |S|$) depending on the minimum amount of members that should be in the clusters ($|C_{min}|$) before they are accepted. This parameter is less important when applied on small samples, but was kept as a variable in the proposed persona development method to accommodate persona development based on larger data sets where a more fine-grained control of the minimum cluster size may be needed.

Both PDs appeared to have a different strategy when it came to group sizes in their manual approaches. PD1 grouped interviewees consistently with two to four members in each group, while PD2 had two large groups of seven members and two smaller with four members. While the DOC algorithm in this paper was run with four different $\beta$ settings, the PD applying the DOC to create personas would normally select one value for the $\beta$ parameter and evaluate the produced subspace clusters. Setting the $\beta$ parameter enforced a deliberate choice on how many samples should be used to define a subspace, e.g. if broad patterns or local characteristics should be found. We recommend setting the $\beta$ parameter high, inspecting the output proto-personas and then adjusting the parameter until a satisfactory level of detail has been reached for the proto-personas. It should be noted that \cite{Procopiuc2002} sets an upper limit of 0.5 for the beta-parameter to ensure a 2-approximate solution (see Theorem 1). We recommend adhering to this limit when using larger datasets where execution times become of greater relevance. On smaller datasets, a $\beta$ parameter higher than 0.5 can be applied and the algorithm run several times to check the stability of the clusters. To settle on a value for $\beta$, we found it useful to consider and determine the size of the discrimination set $r$ (how many interviewees are used to find the subspace) and then set the value of $\beta$ to reflect this, for instance by using the formula $ \beta = 2e^{-ln(2|d|)/r} $.

%The DOC algorithm is only able to find subspace clusters that have sizes equal to or above the size of the discrimination set. Setting the $\beta$ parameter high will 

\subsection{Handling similar subspace clusters}
\label{sec:highmember}
%\subsubsection{Proto-personas by using individual subspace clusters}
Each individual subspace cluster can serve as a proto-persona. For example, subspace cluster O65 in Table~\ref{tab:large65subspace} covered behaviours shared among participants with IDs 12, 13, 14, and 17, and it can be represented by the following proto-persona description.

\textit{Anti-tech grandmother: She} (d47) \textit{lives alone} (d46) \textit{and, although she struggles with some health problems,} (d1) \textit{she manages to eat out several times a week,} (d15) \textit{usually at the elderly caf\'e} (d16). \textit{She spends most of her time by herself but engages in social activities from time to time} (d5). \textit{She has good contacts with her family} (d7) \textit{and has little interest in IT and technology} (d12).

The range of subspace clusters allows the PD to choose the subspace clusters that create the most diverse set of users or fit the problem area better while ensuring that the proto-persona will be based on an optimal cluster, created through a rigorous process.

\begin{figure}[h]
\includegraphics[angle=0,scale=0.066,trim={0cm 5cm 0cm 0cm},clip]{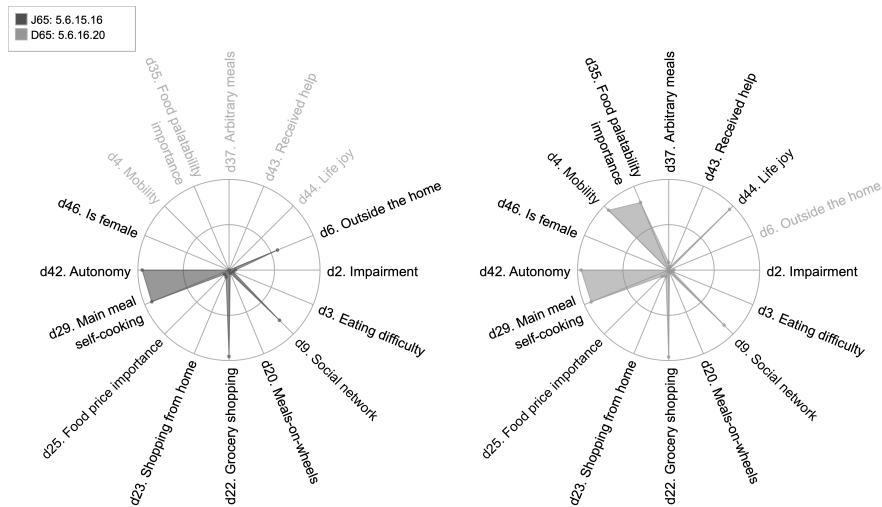}
\caption{Proto-persona radar plots generated from subspace cluster J65 (left) and D65 (right).}
\label{fig:rawsubspacecompare}
\end{figure}

However, as both PDs made use of partial members in their segmentations, we here outline a structured approach on how to merge the subspace clusters that appear very similar. To compare the proto-personas, we represented them as custom radar plots, inspired by the work of Christiernin \citep{Christiernin2010}. The radar plots contained all of the dimensions of the two proto-personas that we compared. Each dimension was labelled with one extreme, and the centre of the radar represented the opposite extreme. The label of a dimension was greyed out if the proto-persona did not cluster in the specific dimension. Comparing the shapes of the radar plots allowed us to distinguish the differences between two proto-personas, and then (if appropriate) be named according to their unique characteristics. Examples of two individual subspace clusters that yielded very similar proto-personas are D65 and J65 (see Figure~\ref{fig:rawsubspacecompare}).

% Figure

The two clusters were made from the same key participants (5, 6 and 16), but each had a unique fourth member (15 and 20) as part of the cluster who did not share all of the dimensions with the core participants. As can be seen from the shape of the visualization of the two proto-personas, they are almost identical, with D65 being an extension of J65. The additional dimensions of D65 are meaningful in relation to the shared dimensions between D65 and J65. As an example, autonomy behaviour (d42) often is in line with receiving little help from others (d43). The unique dimension of J65 does not contradict the additional dimensions of D65.

To simulate the partial membership groupings performed by the PDs we propose calculating a similarity score quantifying similarities between subspace clusters to merge members of the clusters. The score between any two clusters ($(C_a,D_a)$ and $(C_b,D_b)$) might be calculated by finding the shared set $F$ of dimensions:
\begin{equation}
F = D_a \cap D_b
\end{equation}

The vector $m_{(C,D)} = (m_{(C,D)_1},m_{(C,D)_2},\dotsc,m_{(C,D)_{|F|}})$ containing the means of the cluster members on each shared dimension in the subspace can then calculated for each cluster, as follows:
\begin{equation}
m_{(C,D)_j} = \frac{1}{|C|}\sum_{i=1}^{|C|}c_{i,j} \qquad for\: j = 1,\dotsc,|F|
\end{equation}

The similarity score between any two clusters ($(C_{a},D_{a})$ and $(C_{b},D_{b})$) may be obtained using the Dice similarity coefficient \citep{Dice1945a} scaled by the squared euclidean distance divided by the maximum squared distance, which is the same as the number of shared dimensions when dimensions are in the interval between zero and one.
\begin{equation}
\label{eq:similarity_score}
 sim((C_{a},D_{a}),(C_{b},D_{b})) = \frac{2|F|}{|D_{a}|+|D_{b}|}\cdot \frac{\sum_{i=1}^{|F|}(m_{(C_{b},D_{b})_i}-m_{(C_{a},D_{a})_i})^2}{|F|} 
\end{equation}

\begin{figure}[ht]
\includegraphics[angle=-90,scale=0.5,trim={1.5cm 3cm 1.5cm 3},clip]{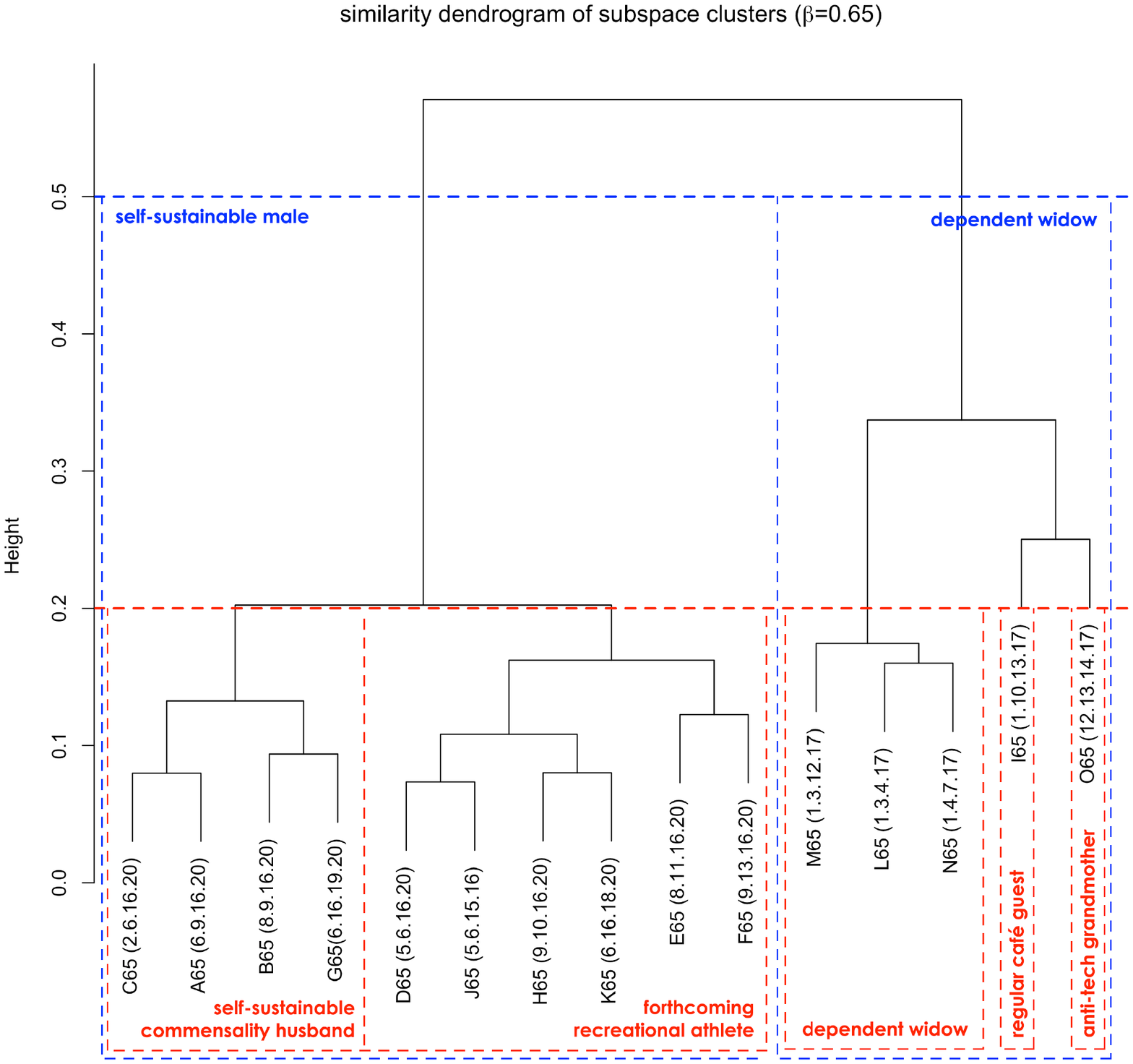}
\caption{Dendrogram obtained with $beta=0.65$ where the blue outline shows the sets of clusters to merge when the dendrogram is cut at $height=0.5$ and the red outline indicates the sets of clusters to merge with a cut at $height=0.2$. The sets are given a name representing the behavior of the proto-persona created from the merged set of clusters.}
\label{fig:large65dendrogram}
\end{figure}

The similarity score ranged between 0 and 1, and it accounts for the number of shared dimensions scaled by the distance between the mean vectors of each cluster. If the subspace clusters have the same means and the same dimensions, the score will be 1. If none of the dimensions are shared and/or the two vectors have means located at opposite extremes of the dimensions, the score will be 0. The similarity score only considers the subspace and the central tendency while member labels are not important for characterizing common behaviour. The matrix of the similarity scores between any combinations of clusters can be used to provide distance measures for agglomerative hierarchical clustering. Cutting the resulting dendrograms at a certain dissimilarity threshold ($1-similarity$ \citep{Podani2000}) will reveal the set of subspace clusters that should be merged. Cutting the dendrogram at a low height will yield many sets comprising a few subspace clusters with minor differences, whereas cutting it at a higher point would produce a few sets of subspaces with high dissimilarity between them (see Figure~\ref{fig:large65dendrogram}).

% Figure

To merge a set of clusters, we only included behavioural dimensions which were part of the subspace of minimum half of the clusters in the set. The arithmetic mean of each dimension (calculated across the clusters in the set containing a non-NA value) was then treated as the dimension score for the proto-persona created from the merged set of clusters if the variance was small and the dimensions were non-conflicting (see Table~\ref{tab:extractathlete} for an example).

\begin{table}[ht]
\centering\scriptsize
\setlength\tabcolsep{2pt} % General space between columns (6pt standard)
\renewcommand{\arraystretch}{1.0} % General space between rows (1 standard)
\caption{Sorted behavioural dimensions that comprised part of the subspace for half or more of the clusters  D65, E65, F65, H65, J65, and K65 in the ``forthcoming recreational athlete'' branch.}
\label{tab:extractathlete}%
%\begin{tabulary}{0.8\linewidth}{LCCCCCCRR}
\begin{tabular}{lccccccrr}
& D65 & E65 & F65 & H65 & J65 & K65 & \multicolumn{1}{|r}{Mean} & \multicolumn{1}{r}{Std. dev.} \\\toprule
Members & \shortstack{5.6.\\16.20} & \shortstack{8.11.\\16.20} & \shortstack{9.13.\\16.20} & \shortstack{9.10.\\16.20} & \shortstack{5.6.\\15.16} & \shortstack{6.16.\\18.20} & \multicolumn{2}{c}{Merged} \\\toprule
d20. Frequency of meals-on-wheels & 0.029 & 0.026 & 0.024 & 0.021 & 0.038 & 0.029 & \multicolumn{1}{|r}{0.028} & 0.006 \\
d3. Eating difficulty & 0.046 & NA & 0.044 & 0.038 & 0.051 & 0.067 & \multicolumn{1}{|r}{0.049} & 0.011 \\
d22. Frequency of grocery shopping & 0.961 & 0.962 & 0.958 & 0.960 & 0.956 & NA & \multicolumn{1}{|r}{0.960} & 0.002 \\
d42. Autonomy & 0.962 & 0.962 & 0.958 & NA & 0.953 & 0.890 & \multicolumn{1}{|r}{0.945} & 0.031 \\
d23. Shopping from home & 0.078 & NA & 0.042 & 0.038 & 0.083 & NA & \multicolumn{1}{|r}{0.060} & 0.024 \\
d37. Tendency of arbitrary meals & 0.084 & 0.114 & 0.094 & 0.112 & NA & NA & \multicolumn{1}{|r}{0.101} & 0.015 \\
d47. Is female & 0.000 & NA & NA & 0.000 & 0.000 & 0.000 & \multicolumn{1}{|r}{0.000} & 0.000 \\
d2. Impairment & 0.058 & NA & 0.123 & NA & 0.072 & NA & \multicolumn{1}{|r}{0.084} & 0.034 \\
d4. Mobility & 0.935 & 0.961 & 0.959 & NA & NA & NA & \multicolumn{1}{|r}{0.952} & 0.015 \\
d5. Amount of social activities & NA & 0.925 & NA & 0.911 & NA & 0.923 & \multicolumn{1}{|r}{0.920} & 0.007 \\
d9. Size of social network & 0.861 & NA & NA & NA & 0.785 & 0.804 & \multicolumn{1}{|r}{0.817} & 0.040 \\
d13. Prefers to eat with others & NA & NA & 0.892 & 0.939 & NA & 0.957 & \multicolumn{1}{|r}{0.929} & 0.033 \\
d19. Appeal of eating with new people & NA & 0.740 & 0.752 & 0.819 & NA & NA & \multicolumn{1}{|r}{0.770} & 0.042 \\
d25. Importance of food price & 0.097 & NA & NA & NA & 0.066 & 0.098 & \multicolumn{1}{|r}{0.087} & 0.018 \\
d29. Frequency of cooking main meal & 0.922 & 0.948 & NA & NA & 0.918 & NA & \multicolumn{1}{|r}{0.929} & 0.016 \\
d35. Importance of food sense perception & 0.807 & NA & NA & 0.860 & NA & 0.884 & \multicolumn{1}{|r}{0.850} & 0.040 \\
d41. Amount of physical exercise & NA & 0.916 & 0.796 & 0.827 & NA & NA & \multicolumn{1}{|r}{0.846} & 0.063 \\
d44. Life joy & 0.946 & 0.953 & NA & NA & NA & 0.950 & \multicolumn{1}{|r}{0.949} & 0.004 \\\bottomrule
\end{tabular}%
\end{table}%

\FloatBarrier

\subsection{From proto-personas to final personas}

The proto-personas obtained should be used to generate characteristics and relevant goals in stage four of Goodwin's method. From a selected subspace cluster or a set of merged clusters characteristics can be summarized in descriptions, such as the one below for L45 (Table~\ref{tab:large45subspace}), which we labelled ``The isolated meal-skipping widow.'' 

\textit{Isolated meal-skipping widow: She} (d47) \textit{is living by herself} (d46) \textit{, and, due to health problems} (d1) \textit{and impairments} (d2), \textit{she is highly dependent on others} (d42 and d43) \textit{, and the only time she leaves her home is to shop for necessities} (d6 and d23). \textit{Her smell and taste sensations are lowered} (d34) \textit{, and, although she prefers to eat her meals with others} (d13), \textit{she eats most of her meals alone} (d14) \textit{and rarely engages in social activities} (d5). {Her eating schedule is arbitrary and skipping a meal is not uncommon for her} (d37).

Goodwin suggested that three to four goals should be generated per persona and that ``for each of the proto-personas, there's usually at least one goal evident from the mapping exercise [stage three]'' (Goodwin recommends studying the data obtained from grouped interviewees to determine the remaining goals) \citep{Goodwin2009}. A suitable goal can be described as, ``...something the product [in our case \textit{intervention}] can help people accomplish, but can't entirely accomplish for them'' \citep{Goodwin2009}. For example, according to the description of the isolated meal-skipping widow, she appears to have the goal of eating with others, although her behaviour and situation do not align with this goal. Thus, for example, an internet-based communication solution could be attempted for this persona to help her achieve her goal and increase her motivation to eat more consistently.

When goals are listed for each proto-persona (step 4) and too-similar proto-personas have been merged or eliminated (step 5), the descriptions can be expanded into narratives and paired with a name and representative pictures to create the final personas (step 6).

\begin{sidewaystable}[h]
\renewcommand{\arraystretch}{0.95}
\centering\scriptsize
\setlength\tabcolsep{4pt} % General space between columns (6pt standard)
\caption{Subspace clusters with $\beta=0.25$\vspace*{-6pt}}
\label{tab:smallsubspace}%
\begin{tabulary}{\linewidth}{LCCCCCCCCCCCCCCCCC}
%\begin{tabular}{lccccccccccccccccc}
& A25   & B25   & C25   & D25   & E25   & F25   & G25   & H25   & I25   & J25   & K25   & L25   & M25   & N25   & O25   & P25   & Q25 \\
Members & 16.20 & 6.16  & 5.20  & 9.16  & 8.20  & 12.17 & 1.17  & 11.13 & 2.20  & 1.10  & 5.19  & 5.15  & 3.17  & 1.4   & 4.7   & 10.18 & 10.14 \\\toprule
d1. Healthy & 0.958 & 0.933 & NA    & NA    & NA    & 0.332 & 0.286 & 0.238 & NA    & 0.331 & 0.248 & 0.363 & 0.174 & 0.155 & NA    & 0.316 & 0.390 \\
d2. Impairment & 0.042 & 0.037 & 0.080 & 0.097 & 0.042 & 0.768 & 0.922 & NA    & 0.145 & NA    & NA    & 0.107 & 0.935 & 0.835 & NA    & 0.605 & 0.681 \\
d3. Eating difficulty & 0.037 & 0.050 & 0.042 & 0.032 & NA    & 0.075 & 0.043 & 0.058 & NA    & 0.038 & 0.046 & 0.053 & NA    & NA    & 0.812 & 0.091 & NA \\
d4. Mobility & 0.960 & 0.954 & 0.915 & 0.962 & 0.960 & 0.081 & 0.091 & 0.958 & 0.863 & 0.256 & NA    & 0.764 & 0.113 & NA    & 0.869 & 0.374 & NA \\
d5. Amount of social activities & 0.903 & 0.909 & NA    & 0.896 & 0.946 & 0.313 & 0.224 & NA    & NA    & NA    & NA    & 0.486 & 0.374 & 0.150 & 0.109 & 0.930 & NA \\
d6. Activities outside the home & 0.754 & 0.666 & NA    & 0.570 & 0.925 & 0.147 & 0.281 & 0.613 & NA    & 0.412 & NA    & 0.487 & 0.147 & 0.283 & 0.150 & NA    & 0.551 \\
d7. Family contact & NA    & 0.944 & NA    & NA    & NA    & 0.645 & NA    & 0.869 & NA    & NA    & 0.813 & NA    & NA    & 0.059 & NA    & 0.935 & 0.863 \\
d8. Contact with friends & NA    & NA    & 0.907 & NA    & 0.942 & NA    & 0.355 & NA    & NA    & NA    & NA    & NA    & NA    & 0.511 & NA    & 0.906 & 0.748 \\
d9. Size of social network & 0.807 & 0.804 & 0.919 & NA    & 0.944 & NA    & NA    & NA    & NA    & NA    & NA    & 0.767 & NA    & NA    & 0.284 & NA    & NA \\
d10. IT competences & NA    & NA    & 0.551 & 0.963 & NA    & 0.046 & NA    & 0.131 & 0.591 & 0.540 & 0.518 & 0.633 & 0.083 & NA    & NA    & NA    & 0.575 \\
d11. Appeal of Skype-meal & 0.043 & NA    & 0.045 & NA    & 0.050 & NA    & 0.064 & NA    & NA    & NA    & NA    & NA    & NA    & NA    & NA    & NA    & NA \\
d12. IT/Technology interest & NA    & NA    & 0.535 & 0.829 & 0.254 & 0.022 & NA    & 0.096 & 0.494 & 0.564 & NA    & 0.668 & NA    & NA    & NA    & NA    & 0.390 \\
d13. Prefers to eat with others & 0.963 & 0.942 & NA    & 0.923 & 0.960 & 0.859 & 0.899 & 0.625 & 0.962 & 0.939 & NA    & 0.169 & 0.896 & 0.954 & NA    & 0.958 & NA \\
d14. Frequency of meals alone & 0.062 & 0.056 & NA    & 0.051 & 0.097 & 0.877 & 0.903 & 0.514 & 0.065 & NA    & NA    & 0.933 & 0.712 & 0.970 & 0.893 & 0.500 & 0.494 \\
d15. Tendency to eat out & 0.323 & NA    & 0.318 & NA    & NA    & 0.498 & 0.693 & 0.617 & NA    & 0.783 & NA    & NA    & NA    & NA    & 0.040 & 0.666 & 0.681 \\
d16. Tendency to eat at elder cafe & NA    & 0.037 & 0.655 & 0.027 & NA    & 0.923 & 0.966 & NA    & NA    & 0.955 & 0.546 & NA    & NA    & NA    & 0.026 & 0.834 & 0.927 \\
d17. Satisfaction with food at elder cafe & NA    & NA    & 0.931 & NA    & NA    & NA    & NA    & NA    & NA    & 0.962 & 0.866 & NA    & NA    & NA    & NA    & NA    & 0.941 \\
d18. Frequency of restaurant visits & NA    & NA    & NA    & NA    & NA    & NA    & NA    & NA    & NA    & NA    & NA    & NA    & NA    & NA    & NA    & NA    & NA \\
d19. Appeal of eating with new people & 0.799 & NA    & NA    & 0.740 & 0.744 & NA    & NA    & 0.701 & 0.748 & NA    & NA    & 0.038 & 0.061 & NA    & NA    & 0.947 & NA \\
d20. Frequency of meals-on-wheels & 0.021 & 0.029 & 0.029 & 0.019 & 0.026 & NA    & 0.024 & 0.034 & 0.021 & 0.019 & 0.030 & 0.046 & 0.032 & NA    & 0.933 & 0.029 & NA \\
d21. Satisfaction with meals-on-wheels & NA    & NA    & NA    & NA    & NA    & NA    & NA    & NA    & NA    & NA    & NA    & NA    & NA    & NA    & 0.935 & NA    & NA \\
d22. Frequency of grocery shopping & 0.965 & 0.960 & 0.962 & 0.962 & 0.968 & 0.109 & NA    & 0.947 & 0.958 & 0.882 & 0.958 & 0.952 & NA    & 0.887 & NA    & NA    & NA \\
d23. Shopping from home & 0.035 & 0.121 & 0.035 & 0.042 & 0.029 & 0.051 & 0.037 & NA    & NA    & 0.035 & 0.037 & 0.045 & 0.050 & 0.037 & 0.064 & NA    & NA \\
d24. Appeal of delivered groceries & NA    & 0.754 & NA    & 0.633 & NA    & NA    & NA    & 0.939 & NA    & NA    & 0.147 & NA    & 0.067 & NA    & NA    & NA    & NA \\
d25. Importance of food price & 0.109 & 0.085 & 0.110 & NA    & NA    & 0.067 & NA    & NA    & 0.125 & NA    & 0.067 & 0.048 & NA    & NA    & NA    & NA    & NA \\
d26. Importance of food quality & NA    & NA    & NA    & NA    & NA    & NA    & NA    & NA    & NA    & NA    & 0.949 & NA    & NA    & NA    & NA    & NA    & NA \\
d27. Importance of shopping convenience & NA    & NA    & NA    & NA    & NA    & NA    & NA    & NA    & NA    & NA    & NA    & NA    & NA    & NA    & NA    & NA    & NA \\
d28. Importance of ecology & NA    & NA    & NA    & NA    & NA    & NA    & NA    & NA    & NA    & NA    & 0.938 & NA    & NA    & NA    & NA    & NA    & NA \\
d29. Frequency of cooking main meal & 0.947 & 0.904 & 0.939 & NA    & 0.955 & NA    & 0.403 & NA    & 0.952 & 0.359 & 0.925 & 0.931 & 0.473 & NA    & NA    & NA    & NA \\
d30. Cooking enthusiasm & NA    & NA    & NA    & NA    & NA    & NA    & NA    & NA    & NA    & 0.597 & NA    & NA    & NA    & NA    & NA    & 0.589 & NA \\
d31. Satisfaction with kitchen conditions & 0.957 & NA    & 0.963 & NA    & NA    & 0.928 & NA    & 0.133 & 0.930 & NA    & 0.911 & 0.954 & NA    & 0.709 & NA    & NA    & NA \\
d32. Importance of reducing food waste & NA    & NA    & NA    & NA    & NA    & NA    & NA    & NA    & 0.914 & NA    & NA    & NA    & NA    & NA    & NA    & NA    & NA \\
d33. Food for pleasure rather than survival & NA    & 0.904 & 0.486 & 0.957 & 0.484 & 0.165 & 0.157 & 0.946 & 0.377 & NA    & NA    & NA    & 0.152 & NA    & NA    & NA    & NA \\
d34. High smell and taste sense perception & 0.791 & 0.720 & 0.629 & 0.708 & NA    & NA    & 0.152 & NA    & NA    & NA    & 0.597 & NA    & 0.157 & 0.117 & 0.327 & NA    & NA \\
d35. Importance of food sense perception & 0.880 & 0.872 & 0.741 & 0.931 & NA    & NA    & NA    & NA    & 0.837 & 0.698 & 0.778 & NA    & NA    & 0.770 & 0.888 & 0.845 & NA \\
d36. Importance of meal setting & 0.912 & 0.805 & NA    & NA    & 0.819 & NA    & NA    & NA    & 0.861 & 0.716 & NA    & NA    & NA    & NA    & NA    & NA    & NA \\
d37. Tendency to eat arbitrary meals & 0.064 & 0.086 & 0.081 & 0.059 & 0.064 & 0.487 & 0.462 & 0.230 & 0.051 & 0.372 & 0.077 & NA    & NA    & 0.361 & NA    & 0.342 & 0.195 \\
d38. Appetite & 0.767 & 0.609 & 0.863 & 0.815 & 0.772 & NA    & NA    & 0.891 & NA    & 0.059 & 0.923 & NA    & NA    & 0.102 & 0.243 & NA    & 0.046 \\
d39. Importance of food health & 0.911 & 0.899 & NA    & 0.911 & 0.960 & NA    & NA    & 0.879 & 0.954 & NA    & NA    & NA    & NA    & NA    & NA    & NA    & NA \\
d40. Wishes to lose weight & NA    & NA    & NA    & NA    & NA    & NA    & NA    & NA    & NA    & NA    & NA    & NA    & NA    & NA    & NA    & NA    & NA \\
d41. Amount of physical exercise & 0.880 & NA    & 0.843 & 0.815 & 0.898 & 0.061 & 0.048 & 0.831 & NA    & NA    & NA    & NA    & 0.102 & NA    & 0.620 & NA    & 0.882 \\
d42. Autonomy & 0.962 & 0.957 & 0.968 & 0.955 & 0.973 & 0.185 & 0.396 & 0.950 & 0.971 & 0.482 & 0.958 & 0.949 & 0.308 & 0.500 & 0.506 & 0.567 & 0.436 \\
d43. Amount of help received & 0.037 & 0.040 & 0.038 & 0.043 & 0.034 & 0.843 & 0.711 & NA    & NA    & 0.612 & 0.050 & NA    & 0.781 & 0.653 & 0.618 & 0.564 & NA \\
d44. Enjoyment of life & 0.954 & 0.935 & 0.957 & 0.944 & 0.965 & 0.682 & NA    & NA    & 0.963 & NA    & 0.946 & NA    & NA    & 0.134 & 0.326 & NA    & 0.711 \\
d45. Retrospective thinking & NA    & NA    & NA    & NA    & NA    & NA    & NA    & NA    & NA    & NA    & NA    & NA    & NA    & NA    & NA    & NA    & NA \\
d46. Living with partner & 1.000 & 1.000 & NA    & 1.000 & 1.000 & 0.000 & 0.000 & 0.000 & 1.000 & 0.000 & NA    & 0.000 & 0.000 & 0.000 & 0.000 & 0.000 & 0.000 \\
d47. Is female & 0.000 & 0.000 & 0.000 & 0.000 & NA    & 1.000 & 1.000 & 1.000 & 0.000 & NA    & NA    & 0.000 & 1.000 & 1.000 & 1.000 & 0.000 & NA \\
\bottomrule Dimensions & 30.00 & 29.00 & 27.00 & 27.00 & 25.00 & 25.00 & 23.00 & 23.00 & 22.00 & 22.00 & 22.00 & 21.00 & 20.00 & 20.00 & 20.00 & 20.00 & 17.00 \\
\end{tabulary}%
\end{sidewaystable}%

\begin{sidewaystable}[h]
\renewcommand{\arraystretch}{0.95}
\centering\scriptsize
\setlength\tabcolsep{3pt} % General space between columns (6pt standard)
\caption{Subspace clusters with $\beta=0.45$\vspace*{-6pt}}
\label{tab:large45subspace}%
%\begin{tabulary}{\linewidth}{LCCCCCCCCCCCCCCCC}
\begin{tabular}{lcccccccccccccccc}
& A45   & B45   & C45   & D45   & E45   & F45   & G45   & H45   & I45   & J45   & K45   & L45   & M45   & N45   & O45   & P45 \\
Members & 6.16.20 & 6.9.16 & 8.16.20 & 5.16.20 & 2.16.20 & 11.16.20 & 6.13.16 & 5.19.20 & 1.3.17 & 1.12.17 & 6.16.18 & 1.4.17 & 5.10.20 & 11.13.14 & 5.15.16 & 7.12.17 \\\toprule
d1. & 0.940 & NA    & NA    & NA    & NA    & NA    & NA    & NA    & 0.210 & 0.315 & NA    & 0.200 & NA    & NA    & NA    & NA \\
d2. & 0.040 & 0.078 & 0.039 & 0.065 & 0.109 & NA    & 0.108 & NA    & 0.933 & NA    & NA    & 0.862 & NA    & NA    & 0.083 & NA \\
d3. & 0.046 & 0.043 & NA    & 0.039 & NA    & 0.038 & 0.058 & 0.044 & NA    & 0.059 & 0.077 & NA    & 0.045 & NA    & 0.047 & NA \\
d4. & 0.954 & 0.955 & 0.962 & 0.932 & 0.897 & 0.960 & 0.955 & NA    & 0.118 & 0.097 & NA    & NA    & NA    & 0.944 & NA    & NA \\
d5. & 0.917 & 0.913 & 0.921 & NA    & NA    & 0.914 & NA    & NA    & NA    & 0.267 & 0.919 & 0.191 & NA    & NA    & NA    & 0.240 \\
d6. & 0.742 & 0.620 & NA    & NA    & NA    & 0.750 & 0.606 & NA    & 0.210 & 0.210 & NA    & 0.264 & NA    & 0.613 & 0.529 & 0.121 \\
d7. & NA    & NA    & NA    & NA    & NA    & NA    & 0.905 & NA    & NA    & NA    & 0.940 & NA    & NA    & 0.842 & NA    & 0.642 \\
d8. & NA    & NA    & NA    & NA    & NA    & NA    & NA    & NA    & NA    & NA    & NA    & NA    & 0.903 & NA    & NA    & NA \\
d9. & 0.843 & NA    & 0.859 & 0.842 & NA    & NA    & NA    & NA    & NA    & NA    & 0.764 & NA    & NA    & NA    & 0.741 & NA \\
d10. & NA    & NA    & NA    & NA    & NA    & NA    & NA    & 0.527 & NA    & NA    & NA    & NA    & 0.564 & NA    & NA    & 0.044 \\
d11. & NA    & NA    & 0.038 & 0.035 & NA    & 0.050 & NA    & NA    & NA    & NA    & NA    & NA    & NA    & NA    & NA    & NA \\
d12. & NA    & NA    & NA    & NA    & NA    & NA    & NA    & NA    & NA    & NA    & NA    & NA    & 0.523 & 0.158 & NA    & 0.029 \\
d13. & 0.951 & 0.924 & 0.960 & NA    & 0.961 & NA    & 0.880 & NA    & 0.909 & 0.885 & 0.953 & 0.923 & NA    & 0.576 & NA    & NA \\
d14. & 0.063 & 0.055 & 0.081 & NA    & 0.060 & NA    & NA    & NA    & NA    & 0.902 & NA    & 0.931 & NA    & 0.511 & NA    & 0.851 \\
d15. & NA    & NA    & NA    & 0.274 & NA    & NA    & NA    & NA    & NA    & NA    & NA    & NA    & NA    & 0.609 & NA    & NA \\
d16. & NA    & 0.036 & NA    & NA    & NA    & NA    & NA    & 0.617 & NA    & 0.944 & NA    & NA    & NA    & NA    & NA    & NA \\
d17. & NA    & NA    & NA    & NA    & NA    & NA    & NA    & 0.891 & NA    & NA    & NA    & NA    & 0.941 & NA    & NA    & NA \\
d18. & NA    & NA    & NA    & NA    & NA    & NA    & NA    & NA    & NA    & NA    & NA    & NA    & NA    & NA    & NA    & NA \\
d19. & NA    & NA    & 0.747 & NA    & 0.749 & 0.772 & NA    & NA    & NA    & NA    & NA    & NA    & NA    & NA    & NA    & NA \\
d20. & 0.027 & 0.026 & 0.023 & 0.026 & 0.020 & 0.024 & 0.031 & 0.028 & 0.027 & NA    & 0.031 & NA    & 0.027 & NA    & 0.037 & NA \\
d21. & NA    & NA    & NA    & NA    & NA    & NA    & NA    & NA    & NA    & NA    & NA    & NA    & NA    & NA    & NA    & NA \\
d22. & 0.962 & 0.960 & 0.967 & 0.963 & 0.961 & 0.960 & 0.955 & 0.961 & NA    & NA    & NA    & NA    & 0.958 & NA    & 0.956 & 0.097 \\
d23. & 0.092 & 0.096 & 0.032 & 0.036 & NA    & NA    & 0.098 & 0.035 & 0.044 & 0.045 & NA    & 0.038 & 0.036 & NA    & 0.043 & 0.063 \\
d24. & NA    & 0.671 & NA    & NA    & NA    & NA    & 0.808 & NA    & NA    & NA    & 0.808 & NA    & NA    & 0.932 & NA    & NA \\
d25. & 0.121 & NA    & NA    & 0.081 & 0.091 & NA    & 0.114 & 0.110 & NA    & NA    & 0.066 & NA    & NA    & NA    & 0.039 & NA \\
d26. & NA    & NA    & NA    & NA    & NA    & NA    & NA    & NA    & NA    & NA    & NA    & NA    & NA    & NA    & NA    & NA \\
d27. & NA    & NA    & NA    & NA    & NA    & NA    & NA    & NA    & NA    & NA    & NA    & NA    & NA    & NA    & NA    & NA \\
d28. & NA    & NA    & NA    & NA    & NA    & NA    & NA    & NA    & NA    & NA    & NA    & NA    & NA    & NA    & NA    & NA \\
d29. & 0.921 & NA    & 0.950 & 0.939 & 0.948 & 0.946 & NA    & 0.935 & 0.428 & NA    & NA    & NA    & NA    & NA    & 0.934 & NA \\
d30. & NA    & NA    & NA    & NA    & NA    & NA    & NA    & NA    & NA    & NA    & NA    & NA    & NA    & NA    & NA    & NA \\
d31. & NA    & NA    & NA    & 0.958 & 0.936 & NA    & NA    & 0.929 & NA    & NA    & NA    & NA    & NA    & NA    & 0.952 & NA \\
d32. & NA    & NA    & NA    & NA    & NA    & NA    & NA    & NA    & NA    & NA    & NA    & NA    & NA    & NA    & NA    & NA \\
d33. & NA    & 0.923 & NA    & NA    & NA    & NA    & 0.911 & NA    & 0.119 & 0.128 & 0.911 & NA    & 0.483 & NA    & NA    & 0.118 \\
d34. & 0.738 & 0.683 & NA    & NA    & NA    & NA    & 0.688 & 0.656 & 0.121 & NA    & 0.678 & 0.162 & NA    & NA    & NA    & NA \\
d35. & 0.862 & 0.896 & NA    & 0.801 & 0.865 & 0.907 & NA    & 0.799 & NA    & NA    & 0.899 & NA    & 0.740 & NA    & NA    & NA \\
d36. & 0.850 & NA    & 0.841 & NA    & 0.869 & NA    & 0.854 & NA    & NA    & NA    & NA    & NA    & NA    & NA    & NA    & NA \\
d37. & 0.079 & 0.076 & 0.064 & 0.076 & 0.055 & 0.131 & 0.122 & 0.072 & NA    & 0.485 & NA    & 0.389 & 0.143 & 0.195 & NA    & NA \\
d38. & NA    & NA    & 0.750 & 0.810 & NA    & 0.831 & NA    & 0.891 & NA    & NA    & NA    & NA    & NA    & NA    & NA    & NA \\
d39. & 0.916 & 0.916 & 0.931 & NA    & 0.927 & 0.907 & 0.885 & NA    & NA    & NA    & NA    & NA    & NA    & NA    & NA    & NA \\
d40. & NA    & NA    & NA    & NA    & NA    & NA    & NA    & NA    & NA    & NA    & NA    & NA    & NA    & NA    & NA    & NA \\
d41. & NA    & NA    & 0.909 & 0.873 & NA    & 0.899 & NA    & NA    & 0.082 & 0.054 & NA    & NA    & 0.846 & 0.858 & NA    & NA \\
d42. & 0.962 & 0.957 & 0.966 & 0.963 & 0.965 & 0.958 & 0.954 & 0.963 & 0.374 & NA    & 0.863 & 0.429 & NA    & NA    & 0.950 & NA \\
d43. & 0.039 & 0.044 & 0.034 & 0.037 & NA    & 0.044 & NA    & 0.046 & 0.749 & 0.790 & NA    & 0.682 & NA    & NA    & NA    & NA \\
d44. & 0.944 & 0.937 & 0.958 & 0.953 & 0.957 & 0.948 & NA    & 0.951 & NA    & NA    & 0.946 & NA    & NA    & NA    & NA    & NA \\
d45. & NA    & NA    & NA    & NA    & NA    & NA    & NA    & NA    & NA    & NA    & NA    & NA    & NA    & NA    & NA    & NA \\
d46. & 1.000 & 1.000 & 1.000 & NA    & 1.000 & NA    & NA    & NA    & 0.000 & 0.000 & NA    & 0.000 & NA    & 0.000 & NA    & 0.000 \\
d47. & 0.000 & 0.000 & NA    & 0.000 & 0.000 & NA    & NA    & NA    & 1.000 & 1.000 & 0.000 & 1.000 & 0.000 & 1.000 & 0.000 & 1.000 \\
\bottomrule Dimensions & 24.00 & 22.00 & 21.00 & 20.00 & 18.00 & 17.00 & 17.00 & 17.00 & 15.00 & 15.00 & 14.00 & 13.00 & 13.00 & 12.00 & 12.00 & 11.00 \\
\end{tabular}%
\end{sidewaystable}%

\begin{sidewaystable}[h]
\renewcommand{\arraystretch}{0.95}
\centering\scriptsize
\setlength\tabcolsep{2pt} % General space between columns (6pt standard)
\caption{Subspace clusters with $\beta=0.65$\vspace*{-6pt}}
\label{tab:large65subspace}%
\begin{tabular}{lccccccccccccccc}
& A65   & B65   & C65   & D65   & E65   & F65   & G65   & H65   & I65   & J65   & K65   & L65   & M65   & N65   & O65 \\
Members & \shortstack{6.9.\\16.20} & \shortstack{8.9.\\16.20} & \shortstack{2.6.\\16.20} & \shortstack{5.6.\\16.20} & \shortstack{8.11.\\16.20} & \shortstack{9.13.\\16.20} & \shortstack{6.16.\\19.20} & \shortstack{9.10.\\16.20} & \shortstack{1.10.\\13.17} & \shortstack{5.6.\\15.16} & \shortstack{6.16.\\18.20} & \shortstack{1.3.\\4.17} & \shortstack{1.3.\\12.17} & \shortstack{1.4.\\7.17} & \shortstack{12.13.\\14.17} \\\toprule
d1. & NA    & NA    & NA    & NA    & NA    & NA    & NA    & NA    & 0.335 & NA    & NA    & 0.165 & NA    & NA    & 0.363 \\
d2. & 0.070 & 0.069 & 0.091 & 0.058 & NA    & 0.123 & NA    & NA    & NA    & 0.072 & NA    & 0.885 & NA    & NA    & NA \\
d3. & 0.042 & NA    & NA    & 0.046 & NA    & 0.044 & 0.046 & 0.038 & 0.053 & 0.051 & 0.067 & NA    & NA    & NA    & NA \\
d4. & 0.955 & 0.961 & 0.908 & 0.935 & 0.961 & 0.959 & NA    & NA    & NA    & NA    & NA    & NA    & 0.116 & NA    & NA \\
d5. & 0.918 & 0.921 & NA    & NA    & 0.925 & NA    & NA    & 0.911 & NA    & NA    & 0.923 & NA    & NA    & 0.166 & 0.414 \\
d6. & NA    & NA    & NA    & NA    & NA    & NA    & NA    & NA    & 0.384 & 0.577 & NA    & 0.215 & 0.174 & 0.216 & NA \\
d7. & NA    & NA    & NA    & NA    & NA    & NA    & NA    & NA    & NA    & NA    & NA    & NA    & NA    & NA    & 0.727 \\
d8. & NA    & NA    & NA    & NA    & NA    & NA    & NA    & NA    & NA    & NA    & NA    & NA    & NA    & NA    & NA \\
d9. & NA    & NA    & NA    & 0.861 & NA    & NA    & NA    & NA    & NA    & 0.785 & 0.804 & NA    & NA    & NA    & NA \\
d10. & NA    & NA    & NA    & NA    & NA    & NA    & NA    & NA    & NA    & NA    & NA    & NA    & NA    & NA    & NA \\
d11. & NA    & NA    & NA    & NA    & 0.045 & NA    & NA    & NA    & NA    & NA    & NA    & NA    & NA    & NA    & NA \\
d12. & NA    & NA    & NA    & NA    & NA    & NA    & NA    & NA    & NA    & NA    & NA    & NA    & NA    & NA    & 0.089 \\
d13. & 0.935 & 0.942 & 0.952 & NA    & NA    & 0.892 & NA    & 0.939 & 0.874 & NA    & 0.957 & 0.925 & 0.896 & NA    & NA \\
d14. & 0.061 & 0.074 & 0.061 & NA    & NA    & NA    & 0.058 & NA    & NA    & NA    & NA    & NA    & NA    & 0.898 & NA \\
d15. & NA    & NA    & NA    & NA    & NA    & NA    & NA    & NA    & 0.694 & NA    & NA    & NA    & NA    & NA    & 0.554 \\
d16. & NA    & NA    & NA    & NA    & NA    & NA    & NA    & NA    & 0.915 & NA    & NA    & NA    & NA    & NA    & 0.894 \\
d17. & NA    & NA    & NA    & NA    & NA    & NA    & NA    & NA    & NA    & NA    & NA    & NA    & NA    & NA    & NA \\
d18. & NA    & NA    & NA    & NA    & NA    & NA    & NA    & NA    & NA    & NA    & NA    & NA    & NA    & NA    & NA \\
d19. & NA    & 0.742 & NA    & NA    & 0.740 & 0.752 & NA    & 0.819 & NA    & NA    & NA    & NA    & NA    & NA    & NA \\
d20. & 0.025 & 0.022 & 0.025 & 0.029 & 0.026 & 0.024 & 0.026 & 0.021 & 0.026 & 0.038 & 0.029 & NA    & NA    & NA    & NA \\
d21. & NA    & NA    & NA    & NA    & NA    & NA    & NA    & NA    & NA    & NA    & NA    & NA    & NA    & NA    & NA \\
d22. & 0.961 & 0.965 & 0.959 & 0.961 & 0.962 & 0.958 & 0.961 & 0.960 & NA    & 0.956 & NA    & NA    & NA    & NA    & NA \\
d23. & 0.080 & 0.035 & NA    & 0.078 & NA    & 0.042 & 0.077 & 0.038 & 0.041 & 0.083 & NA    & 0.043 & 0.048 & 0.050 & NA \\
d24. & NA    & NA    & NA    & NA    & NA    & NA    & NA    & NA    & NA    & NA    & NA    & NA    & NA    & NA    & NA \\
d25. & NA    & NA    & 0.105 & 0.097 & NA    & NA    & 0.118 & NA    & NA    & 0.066 & 0.098 & NA    & NA    & NA    & NA \\
d26. & NA    & NA    & NA    & NA    & NA    & NA    & NA    & NA    & NA    & NA    & NA    & NA    & NA    & NA    & NA \\
d27. & NA    & NA    & NA    & NA    & NA    & NA    & NA    & NA    & NA    & NA    & NA    & NA    & NA    & NA    & NA \\
d28. & NA    & NA    & NA    & NA    & NA    & NA    & NA    & NA    & NA    & NA    & NA    & NA    & NA    & NA    & NA \\
d29. & NA    & NA    & 0.928 & 0.922 & 0.948 & NA    & 0.923 & NA    & 0.417 & 0.918 & NA    & NA    & NA    & NA    & NA \\
d30. & NA    & NA    & NA    & NA    & NA    & NA    & NA    & NA    & NA    & NA    & NA    & NA    & NA    & NA    & NA \\
d31. & NA    & NA    & NA    & NA    & NA    & NA    & NA    & NA    & NA    & NA    & NA    & NA    & NA    & NA    & NA \\
d32. & NA    & NA    & NA    & NA    & NA    & NA    & NA    & NA    & NA    & NA    & NA    & NA    & NA    & NA    & NA \\
d33. & NA    & NA    & NA    & NA    & NA    & NA    & NA    & NA    & NA    & NA    & NA    & NA    & 0.107 & NA    & NA \\
d34. & 0.705 & NA    & NA    & NA    & NA    & 0.703 & 0.731 & NA    & NA    & NA    & 0.702 & 0.137 & NA    & NA    & NA \\
d35. & 0.882 & NA    & 0.855 & 0.807 & NA    & NA    & 0.875 & 0.860 & NA    & NA    & 0.884 & NA    & NA    & NA    & NA \\
d36. & NA    & NA    & 0.833 & NA    & NA    & NA    & NA    & NA    & NA    & NA    & NA    & NA    & NA    & NA    & NA \\
d37. & 0.073 & 0.062 & 0.069 & 0.084 & 0.114 & 0.094 & 0.073 & 0.112 & 0.346 & NA    & NA    & NA    & NA    & NA    & NA \\
d38. & NA    & 0.793 & NA    & NA    & 0.802 & 0.820 & NA    & NA    & NA    & NA    & NA    & NA    & NA    & NA    & NA \\
d39. & 0.924 & 0.935 & 0.927 & NA    & 0.923 & 0.907 & NA    & NA    & NA    & NA    & NA    & NA    & NA    & NA    & NA \\
d40. & NA    & NA    & NA    & NA    & NA    & NA    & NA    & NA    & NA    & NA    & NA    & NA    & NA    & NA    & NA \\
d41. & NA    & 0.856 & NA    & NA    & 0.916 & 0.796 & NA    & 0.827 & NA    & NA    & NA    & NA    & 0.078 & NA    & NA \\
d42. & 0.961 & 0.964 & 0.964 & 0.962 & 0.962 & 0.958 & 0.959 & NA    & NA    & 0.953 & 0.890 & 0.404 & NA    & 0.451 & NA \\
d43. & 0.042 & 0.038 & NA    & 0.039 & 0.040 & NA    & 0.045 & NA    & NA    & NA    & NA    & 0.717 & 0.799 & 0.665 & NA \\
d44. & 0.943 & 0.954 & 0.949 & 0.946 & 0.953 & NA    & 0.942 & NA    & NA    & NA    & 0.950 & NA    & NA    & NA    & NA \\
d45. & NA    & NA    & NA    & NA    & NA    & NA    & NA    & NA    & NA    & NA    & NA    & NA    & NA    & NA    & NA \\
d46. & 1.000 & 1.000 & 1.000 & NA    & NA    & NA    & 1.000 & NA    & 0.000 & NA    & NA    & 0.000 & 0.000 & 0.000 & 0.000 \\
d47. & 0.000 & NA    & 0.000 & 0.000 & NA    & NA    & NA    & 0.000 & NA    & 0.000 & 0.000 & 1.000 & 1.000 & 1.000 & 1.000 \\
\bottomrule Dimensions & 18.00 & 17.00 & 16.00 & 15.00 & 14.00 & 14.00 & 14.00 & 11.00 & 11.00 & 11.00 & 11.00 & 10.00 & 9.00  & 8.00  & 8.00 \\
\end{tabular}%
\end{sidewaystable}%

\begin{sidewaystable}[h]
\renewcommand{\arraystretch}{0.94}
\centering\scriptsize
\setlength\tabcolsep{1pt} % General space between columns (6pt standard)
\caption{Subspace clusters with $\beta=0.85$\vspace*{-6pt}}
\label{tab:large85subspace}%
\begin{tabular}{lccccccccccccc}
& A85   & B85   & C85   & D85   & E85   & F85   & G85   & H85   & I85   & J85   & K85   & L85   & M85 \\
Members & \shortstack{6.8.9.16.20} & \shortstack{2.6.8.\\16.20} & \shortstack{6.9.16.\\19.20} & \shortstack{8.9.13.\\16.20} & \shortstack{8.9.11.\\16.20} & \shortstack{2.5.6.\\16.20} & \shortstack{6.9.10.\\16.20} & \shortstack{5.6.15.\\16.20} & \shortstack{6.9.16.\\18.20} & \shortstack{1.3.5.6.8.9.10.\\13.15.16.19.20} & \shortstack{2.5.6.8.9.10.11.\\14.16.18.19.20} & \shortstack{1.2.4.5.6.8.9.10.11.12.\\13.14.16.17.18.19.20} & \shortstack{1.3.4.5.6.7.8.9.10.12.\\13.15.16.17.19.20} \\\toprule
d1.   & NA    & NA    & NA    & NA    & NA    & NA    & NA    & NA    & NA    & NA    & NA    & NA    & NA \\
d2.   & 0.063 & 0.080 & NA    & 0.105 & NA    & 0.095 & NA    & 0.067 & NA    & NA    & NA    & NA    & NA \\
d3.   & NA    & NA    & 0.043 & NA    & NA    & NA    & 0.043 & 0.049 & 0.059 & NA    & NA    & NA    & NA \\
d4.   & 0.957 & 0.919 & NA    & 0.960 & 0.960 & 0.902 & NA    & NA    & NA    & NA    & NA    & NA    & NA \\
d5.   & 0.926 & NA    & NA    & NA    & 0.924 & NA    & 0.918 & NA    & 0.922 & NA    & NA    & NA    & NA \\
d6.   & NA    & NA    & NA    & NA    & NA    & NA    & NA    & NA    & NA    & NA    & NA    & NA    & NA \\
d7.   & NA    & NA    & NA    & NA    & NA    & NA    & NA    & NA    & NA    & NA    & NA    & NA    & NA \\
d8.   & NA    & NA    & NA    & NA    & NA    & NA    & NA    & NA    & NA    & NA    & NA    & NA    & NA \\
d9.   & NA    & NA    & NA    & NA    & NA    & NA    & NA    & NA    & NA    & NA    & NA    & NA    & NA \\
d10.  & NA    & NA    & NA    & NA    & NA    & NA    & NA    & NA    & NA    & NA    & NA    & NA    & NA \\
d11.  & NA    & NA    & NA    & NA    & NA    & NA    & NA    & NA    & NA    & NA    & NA    & NA    & NA \\
d12.  & NA    & NA    & NA    & NA    & NA    & NA    & NA    & NA    & NA    & NA    & NA    & NA    & NA \\
d13.  & 0.939 & 0.952 & NA    & 0.904 & NA    & NA    & 0.937 & NA    & 0.943 & NA    & NA    & NA    & NA \\
d14.  & 0.072 & 0.072 & 0.057 & NA    & NA    & NA    & NA    & NA    & NA    & NA    & NA    & NA    & NA \\
d15.  & NA    & NA    & NA    & NA    & NA    & NA    & NA    & NA    & NA    & NA    & NA    & NA    & NA \\
d16.  & NA    & NA    & NA    & NA    & NA    & NA    & NA    & NA    & NA    & NA    & NA    & NA    & NA \\
d17.  & NA    & NA    & NA    & NA    & NA    & NA    & NA    & NA    & NA    & NA    & NA    & NA    & NA \\
d18.  & NA    & NA    & NA    & NA    & NA    & NA    & NA    & NA    & NA    & NA    & NA    & NA    & NA \\
d19.  & NA    & NA    & NA    & 0.730 & 0.737 & NA    & NA    & NA    & NA    & NA    & NA    & NA    & NA \\
d20.  & 0.026 & 0.026 & 0.025 & 0.025 & 0.024 & 0.027 & 0.024 & 0.035 & 0.027 & 0.029 & NA    & NA    & NA \\
d21.  & NA    & NA    & NA    & NA    & NA    & NA    & NA    & NA    & NA    & NA    & NA    & NA    & NA \\
d22.  & 0.963 & 0.962 & 0.960 & 0.961 & 0.962 & 0.959 & 0.959 & 0.958 & NA    & 0.944 & NA    & NA    & NA \\
d23.  & 0.069 & NA    & 0.071 & 0.038 & NA    & NA    & 0.072 & 0.073 & NA    & 0.054 & NA    & NA    & 0.055 \\
d24.  & NA    & NA    & NA    & NA    & NA    & NA    & NA    & NA    & NA    & NA    & NA    & NA    & NA \\
d25.  & NA    & NA    & NA    & NA    & NA    & 0.089 & NA    & 0.092 & NA    & NA    & NA    & NA    & NA \\
d26.  & NA    & NA    & NA    & NA    & NA    & NA    & NA    & NA    & NA    & NA    & NA    & NA    & NA \\
d27.  & NA    & NA    & NA    & NA    & NA    & NA    & NA    & NA    & NA    & NA    & NA    & NA    & NA \\
d28.  & NA    & NA    & NA    & NA    & NA    & NA    & NA    & NA    & NA    & NA    & NA    & NA    & NA \\
d29.  & NA    & 0.934 & NA    & NA    & NA    & 0.927 & NA    & 0.925 & NA    & NA    & NA    & NA    & NA \\
d30.  & NA    & NA    & NA    & NA    & NA    & NA    & NA    & NA    & NA    & NA    & NA    & NA    & NA \\
d31.  & NA    & NA    & NA    & NA    & NA    & NA    & NA    & NA    & NA    & NA    & NA    & NA    & NA \\
d32.  & NA    & NA    & NA    & NA    & NA    & NA    & NA    & NA    & NA    & NA    & NA    & NA    & NA \\
d33.  & NA    & NA    & NA    & NA    & NA    & NA    & NA    & NA    & NA    & NA    & NA    & NA    & NA \\
d34.  & NA    & NA    & 0.706 & NA    & NA    & NA    & NA    & NA    & 0.683 & NA    & NA    & NA    & NA \\
d35.  & NA    & NA    & 0.888 & NA    & NA    & 0.812 & 0.853 & NA    & 0.896 & NA    & NA    & NA    & NA \\
d36.  & NA    & 0.806 & NA    & NA    & NA    & NA    & NA    & NA    & NA    & NA    & NA    & NA    & NA \\
d37.  & 0.071 & 0.068 & 0.069 & 0.088 & 0.102 & 0.075 & 0.111 & NA    & NA    & NA    & 0.135 & 0.207 & NA \\
d38.  & NA    & NA    & NA    & 0.799 & 0.826 & NA    & NA    & NA    & NA    & NA    & NA    & NA    & NA \\
d39.  & 0.934 & 0.935 & NA    & 0.919 & 0.928 & NA    & NA    & NA    & NA    & NA    & NA    & NA    & NA \\
d40.  & NA    & NA    & NA    & NA    & NA    & NA    & NA    & NA    & NA    & NA    & NA    & NA    & NA \\
d41.  & NA    & NA    & NA    & 0.830 & 0.872 & NA    & NA    & NA    & NA    & NA    & NA    & NA    & NA \\
d42.  & 0.964 & 0.966 & 0.959 & 0.961 & 0.962 & 0.964 & NA    & 0.957 & 0.904 & NA    & NA    & NA    & NA \\
d43.  & 0.040 & NA    & 0.046 & NA    & 0.042 & NA    & NA    & NA    & NA    & NA    & NA    & NA    & NA \\
d44.  & 0.948 & 0.953 & 0.942 & NA    & 0.951 & 0.950 & NA    & NA    & 0.948 & NA    & 0.910 & NA    & NA \\
d45.  & NA    & NA    & NA    & NA    & NA    & NA    & NA    & NA    & NA    & NA    & NA    & NA    & NA \\
d46.  & 1.000 & 1.000 & 1.000 & NA    & NA    & NA    & NA    & NA    & NA    & NA    & NA    & NA    & NA \\
d47.  & NA    & NA    & NA    & NA    & NA    & 0.000 & 0.000 & 0.000 & 0.000 & NA    & NA    & NA    & NA \\
\bottomrule Dimensions & 14.00 & 13.00 & 12.00 & 12.00 & 12.00 & 11.00 & 9.00  & 9.00  & 9.00  & 3.00  & 2.00  & 1.00  & 1.00 \\
\end{tabular}%
\end{sidewaystable}%

%\FloatBarrier

\vspace*{8pt}

%\section{Discussion}

%--- Provide a “big picture” perspective for readers to remind them of the importance of your study ---
%   Summarize the major gap in understanding that your work is attempting to fill. What was the overarching hypothesis?
%   Why is filling this gap important? How will answering this question move the field forward?

%Personas are used to guide the design process when users are unavailable, thus building the personas from data collected from real users may increase the level of confidence designers have about making decisions based on the persona characteristics. 

\subsection{Conclusion}

At present, a PD who wishes to develop personas based on qualitative data from a limited set of interviewees is left with few methodology options. The typical choice is to apply labour-intensive manual techniques, which are criticized for their lack of rigour and subjective bias. Alternatively, one of the few semi-automated methods attempting to address the limitations of the manual techniques can be used, but they require a strict questioning structure during data collection and are either unable to handle or treat missing data as part of the analysis. The current semi-automated methods also rely on dimensionality reduction techniques, which we argue are difficult to convey to lay members of a design team. A mixed-method has been proposed in this study, which uses manual techniques to account for loose questioning structures and overlapping subspace clustering with automatic feature selection to handle missing data and the high number of dimensions in relation to the low amount of samples. To examine the internal validity of the mixed method, the results from the proposed method were empirically compared with groups found by two human PDs and the perceptual map from an MCA, all extracted from the same authentic data set.

The comparison displayed that segmentation done manually will result in different solutions as details in the method applied will differ from PD to PD. The proposed method can find optimal clusters reliably and efficiently and the co-occurrences of cluster members account for the largest variances mapped out by a MCA. Despite the different results obtained across the manual segmentations, similarities were seen in the groups from the PDs and the resulting clusters of the proposed method.

While the result demonstrates its potential in terms of scalability and reliability, the proposed method comes with a set of limitations. First, the method is a mixed-method, which entails that the first part of the method is performed manually in an interpretive manner. This makes the method subject to the same critique as traditional manual techniques where the lack of rigour and subjective bias can be a threat to the validity of the method. Future studies should attempt to derive natural language processing techniques for the analysis of interview transcripts and extract behavioural variables, which are independent of a strict questioning structure or are able to identify the structure from the texts itself \citep{Young2018}. Despite the biases that the manual interpretive part of the proposed method may cause, it also provides the method with some versatility. The data can come in terms of video, audio, text, drawings, or any other form typical to qualitative data \citep{Bjrner2018} as long as the data by a human individual can be represented as VASs compatible with the subspace clustering.

The automated part of the proposed method also has a set of limiting factors. It should be noted that the sampling method employed by the DOC algorithm yields an approximation of the optimal clusters in the data set. More extensive sampling of large data sets will lead to more accurate results. Another considerable limitation is the relatively large set of similar clusters produced by the overlapping DOC subspace clustering. Even though we recommend reducing the set of unnecessary clusters by selecting or merging the clusters based on the proposed similarity score, it would be more convenient if the algorithm could reduce the number of similar clusters automatically in the future. Therefore, instead of checking for redundancy, the DOC algorithm could check for a high similarity between subspace clusters using the similarity score (line 7 in ALGORITHM~\ref{alg:one}). We were also concerned about the optimal clustering criterion built into the DOC subspace clustering algorithm. This criterion is based on the number of dimensions found in the subspace of the clusters, so it is vulnerable to variables with redundant information. For instance, we included many dimensions related to the social activity (e.g., the ``amount of social activities'' would have correlations with ``family contact'' and ``contact with friends''). By contrast, only one dimension was included related to the area of food waste. Now, we may consider two interviewees, A and B, with an active social life (three dimensions in common) and a third interviewee C with a less active social life. However, interviewee C tends to avoid food waste similarly to interviewee A (one dimension in common), whereas interviewee B pays less attention to food waste. In this case, A and B are likely to be clustered together because the algorithm favours subspace clusters with more dimensions. Also, the clustering algorithm did not return the last two groups selected by the PD because they did not satisfy the optimal cluster criterion. The PD chose to include these clusters based on a single dimension that made the cluster stand out compared with other clusters. Future research could address these issues by investigating alternatives to the optimal cluster criterion, for example, by experimenting with the criterion to yield ``interesting'' subspace clusters that might maximize an orthogonality or dissimilarity score to obtain more diverse user descriptions. In line of thought with the use of extreme personas \citep{Djajadiningrat2000}, a criterion with orthogonal-capabilities would allow the method to create personas that define the outer rim of the user space, and dealing with the needs of these type of personas is likely to also address the needs of less extreme users.

Looking at the method in a broader perspective, there is a common understanding of persona as a description of a fictitious person, but there are no clear definitions, data foundations, or agreements on the benefits of the method. This study is founded within the Cooperian persona perspective, which is described as the goal-directed perspective \citep{Nielsen2002,Floyd2008a}. The core of the goal-directed persona perspective is to develop hypothetical archetypes that are not described as an average person but rather as a unique character with specific details and most often developed using qualitative methods and in the early stages of a software project. However, the method may offer inspiration or be applicable to any persona development method (for example \citep{Pruitt2003}), which, as part of developing personas, seeks patterns in qualitative data or data characterized by high-dimensional sparse data.

While at its present state the proposed method is not fully automated, the current level of automation is an important step forward for persona-based approaches to the design of adaptive experiences such as the works by \cite{Madureira2014}. While beyond the scope of this study, future improvements in this direction could include the investigation of automated extraction of the features from the qualitative data and a study on the relationship between the user personas and their behavioural variables within the adaptive system. The latter would be valuable to connect the personas to real users of the system and enable the correct triggering of the personalised experiences. 

%viability
%The paper has presented an argument of why the proposed method should be easier to convey to layman members.

%generalizability
%Why other clustering methods are not gonna cut it... need to explicitly state in the discussion why subspace clustering is superior compared to other methods!
%Something about the subspace clustering offering a set of unique benefits to handle low sample size data - this could also be beneficial in other types of persona development - ? ? ?

%The alpha parameter.. when would this be relevant??? It sets the minimum size for an acceptable cluster...

\begin{acknowledgements}

This study was part of the ELDORADO project ``Preventing malnourishment and promoting well-being among the older adults at home through personalised cost-effective food and meal supply'' supported by a grant (4105-00009B) from the Innovation Fund Denmark.

\end{acknowledgements}

\FloatBarrier

\bibliographystyle{spbasic}      % basic style, author-year citations
\bibliography{paolo,references}   % name your BibTeX data base

%\begin{thebibliography}{}
% etc
%\end{thebibliography}
\vspace{5mm}
\noindent\textbf{Dannie Korsgaard} is a Data Scientist working at Moodagent a Danish music streaming service and graduated PhD student from the Department of Architecture, Design and Media Technology at Aalborg University. His work is mainly focused on user data management and user modelling to improve recommendations and the user experience with a product.

\vspace{5mm}
\noindent\textbf{Thomas Bjørner} is an Associate Professor at the Department of Architecture, Design and Media Technology at Aalborg University. He is interested in user evaluations and qualitative research advancements, which also imply mixed method strategies. His research addresses besides methodological issues within user evaluations, technology acceptance, interviews, observations, and field studies/ethnography also the characterization of technology usages in different contexts. He is European Commission appointed as an expert within user evaluations and qualitative studies.

\vspace{5mm}
\noindent\textbf{Pernille K. Sørensen} has a Master degree in Educational Anthropology from Aarhus University. Her research focuses mainly on the social and cultural aspects of eating together in various settings, and her research is particularly based on ethnographic methods. In her latest ethnographic fieldwork, she examines a group of drug users’ attachment to a drop-in centre. Her focus is on how the drop-in centre affects drug users in terms of their social life and the practice of eating together.

\vspace{5mm}
\noindent\textbf{Paolo Burelli} is an Assistant Professor at the IT-University of Copenhagen. His research work focuses on using machine learning to model and predict user behaviour in games, with applications to user experience adaptation and business analytics. Dr Burelli has published several articles in journals and conference within the fields of data science, user modelling and artificial intelligence, computer graphics and games. He is currently a member of the IEEE Computational Intelligence Society Games Technical Committee.

\end{document}